\documentclass[10pt,twocolumn,letterpaper]{article}

\usepackage[pagenumbers]{cvpr} 

\usepackage{graphicx}
\usepackage{amsmath}
\usepackage{amssymb}
\usepackage{booktabs}
\usepackage{multirow}
\usepackage{makecell}
\usepackage{bbding}
\usepackage{tablefootnote}
\usepackage{threeparttable}
\usepackage{color}
\definecolor{gray}{RGB}{130, 130, 130}

\usepackage[table]{xcolor}
\usepackage[british, english, american]{babel}
\usepackage[pagebackref=false, breaklinks=true, letterpaper=true, colorlinks,
citecolor=citecolor, linkcolor=linkcolor, bookmarks=false]{hyperref}
\definecolor{citecolor}{HTML}{0071BC}
\definecolor{linkcolor}{HTML}{ED1C24}

\begin{document}
	
	\title{UMIFormer: Mining the Correlations between Similar Tokens \\ for Multi-View 3D Reconstruction}
	
	\author{
		Zhenwei Zhu\footnotemark[2] \hspace{0.05in} Liying Yang\footnotemark[2] \hspace{0.05in} Ning Li \hspace{0.01in} Chaohao Jiang \hspace{0.01in} Yanyan Liang\footnotemark[1] \\
		{Macau University of Science and Technology} \\ \vspace{-5mm}
	}
	
	\maketitle
	
	\renewcommand{\thefootnote}{\fnsymbol{footnote}}
	\footnotetext[2]{Equal contribution. Email: \{garyzhu1996, lyyang69\}@gmail.com}
	\footnotetext[1]{Corresponding author. Email: yyliang@must.edu.mo}
	
	\begin{abstract}
		In recent years, many video tasks have achieved breakthroughs by utilizing the vision transformer and establishing spatial-temporal decoupling for feature extraction. Although multi-view 3D reconstruction also faces multiple images as input, it cannot immediately inherit their success due to completely ambiguous associations between unstructured views. There is not usable prior relationship, which is similar to the temporally-coherence property in a video. To solve this problem, we propose a novel transformer network for Unstructured Multiple Images (UMIFormer). It exploits transformer blocks for decoupled intra-view encoding and designed blocks for token rectification that mine the correlation between similar tokens from different views to achieve decoupled inter-view encoding. Afterward, all tokens acquired from various branches are compressed into a fixed-size compact representation while preserving rich information for reconstruction by leveraging the similarities between tokens. We empirically demonstrate on ShapeNet and confirm that our decoupled learning method is adaptable for unstructured multiple images. Meanwhile, the experiments also verify our model outperforms existing SOTA methods by a large margin. Code will be available at \url{https://github.com/GaryZhu1996/UMIFormer}.
	\end{abstract}
	
	\section{Introduction}
	3D reconstruction, which lifts 2D view images to a 3D representation of an object, is a challenging problem. It plays an important role in numerous technologies, including intelligent driving, augmented reality and robotics. In the situation of single-view input, previous works have attempted to improve performance by strengthening the network capabilities \cite{girdhar2016learning, wu2016learning, smith2017improved, liu2022spatial, xing2022semi} and leveraging the geometric information as priors knowledge \cite{yang2021single, yang2022exploring, xing2022few}. However, for multi-view reconstruction, researchers concentrate on how to extract sufficient feature representation for the object shape from unstructured multiple images \cite{choy20163d, xie2019pix2vox, yang2020robust, xie2020pix2vox++, wang2021multi}. This paper is devoted to multi-view 3D reconstruction using the voxel representation.
	
	\begin{figure}
		\centering
		\begin{subfigure}{0.48\linewidth}
			\scalebox{1.25}{
				\includegraphics[width=0.7\linewidth]{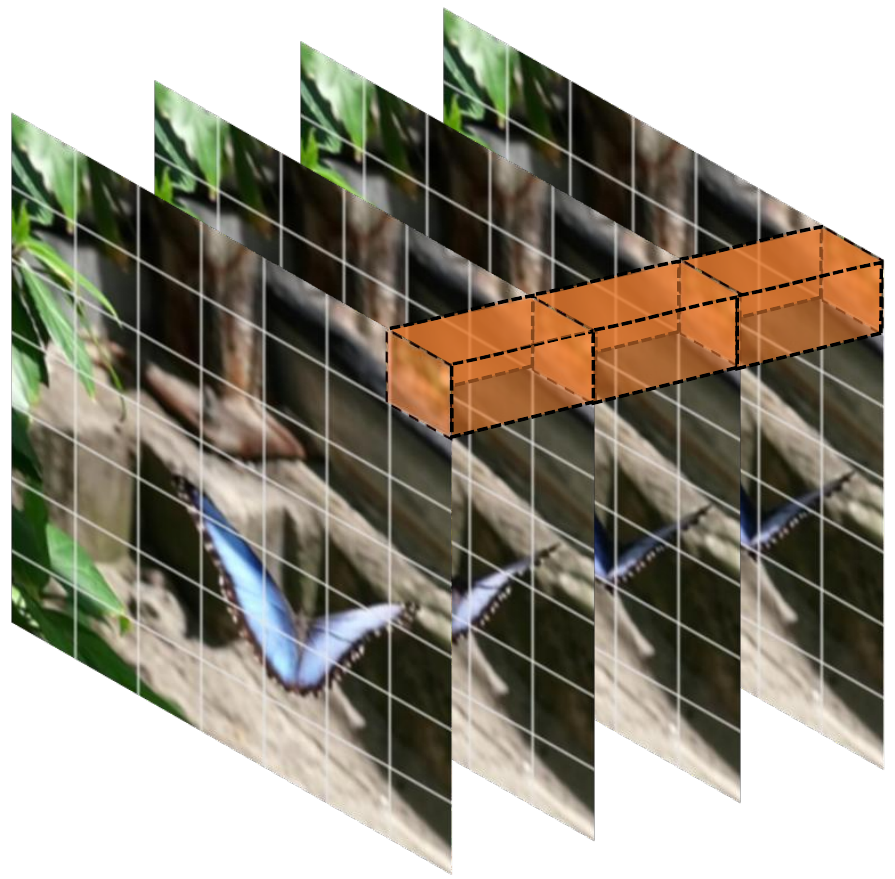}}
			\caption{Video tasks}
			\label{highlight_video}
		\end{subfigure}
		\hfill
		\begin{subfigure}{0.5\linewidth}
			\scalebox{1.25}{
				\includegraphics[width=0.8\linewidth]{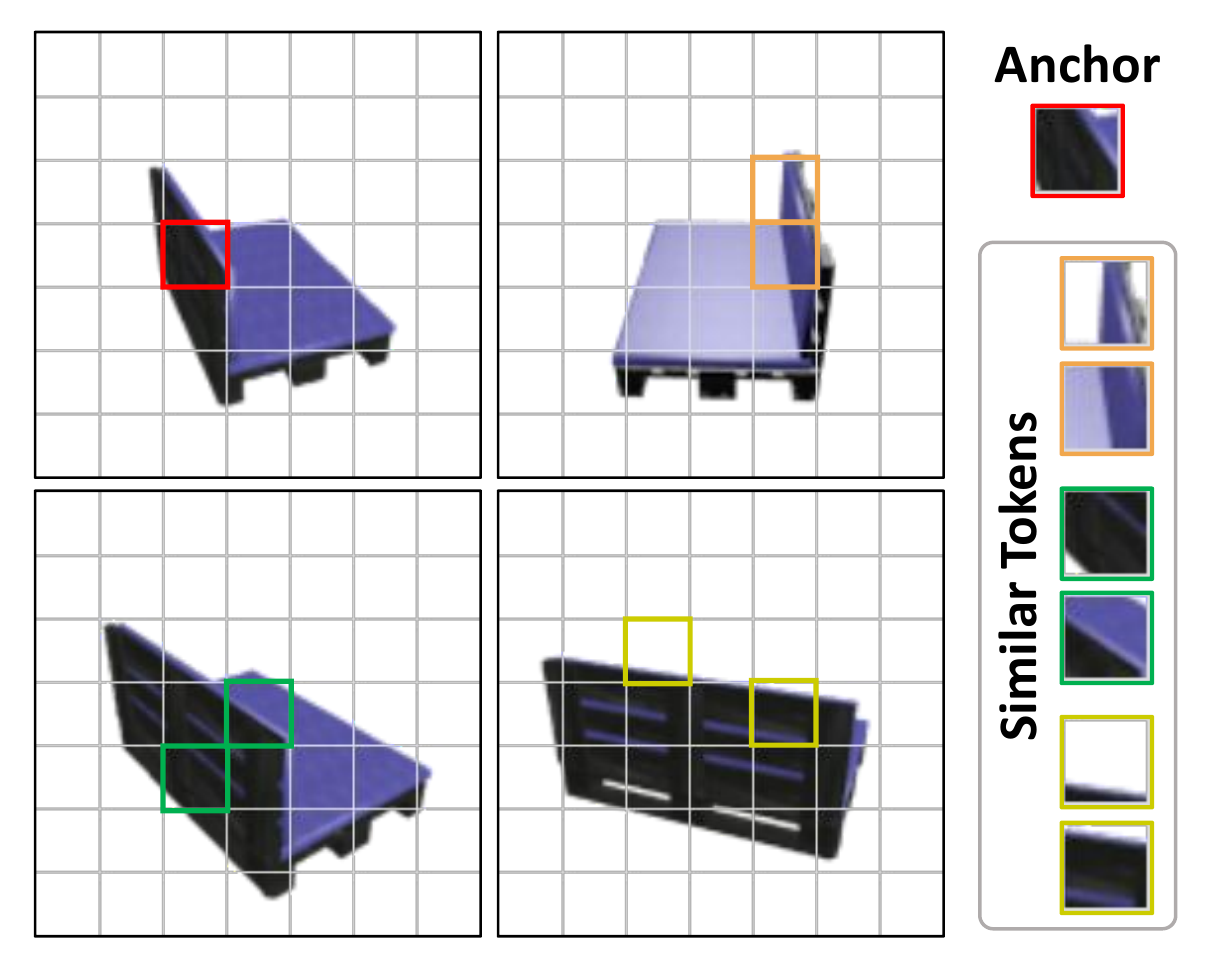}}
			\caption{Multi-view 3D reconstruction}
			\label{highlight_object}
		\end{subfigure}
		\caption{Comparison of the positional correspondence used for inter-image-decoupled feature extraction in (a) video tasks and (b) multi-view 3D reconstruction. Video tasks exploit the temporally-coherence property to establish the prior relationship as shown in (a). For multi-view reconstruction, each patch is treated as an anchor and associated with its similar tokens from other views to build the positional correspondence as shown in (b).}
		\label{highlight}
	\end{figure}
	
	In our investigation, the deep-learning-based algorithms for multi-view reconstruction typically involve two steps: feature extraction and shape reconstruction. The latter is generally accomplished using a 3D decoder module, while there are various solutions for the former including CNN-based and transformer-based methods.
	
	CNN-based methods usually separate the feature extraction process into two stages. The first stage exploits a backbone network to encode on intra-view-dimension while the second stage processing on inter-view-dimension aggregate the features obtained from different views. The fusion method can be a pooling layer \cite{su2015multi, paschalidou2018raynet, huang2018deepmvs}, a recurrent unit \cite{choy20163d, kar2017learning, ma2020improved} or an attention operator \cite{yang2020robust}. In addition, Pix2Vox series \cite{xie2019pix2vox, xie2020pix2vox++} put merger after decoder, which directly fusion the voxels predicted from different views, and also achieve good results. To realize the merger adapting the global state, GARNet \cite{zhu2023garnet} sets up two fusion modules, which are located before and after the decoder respectively.
	
	Transformer-based methods \cite{wang2021multi, yagubbayli2021legoformer, tiong20223d} that can directly handle views as a sequence also attend global awareness. It takes natural advantage of the architecture to couple the procedures for intra-image and inter-image feature extraction. However, such approaches work poorly when facing few views as input since the size of the extracted feature is too small to hold enough information. In contrast, 3D-RETR \cite{shi20213d} exploits transformer on the intra-view dimension and then aggregates the features from different views using an adaptive pooling layer. It is essentially the same method as before \cite{su2015multi} but with a more advanced backbone network. The success of this method reminds us that the power of vision transformer (ViT) \cite{dosovitskiy2021image} for the representation of views cannot be understated.
	
	In video tasks that also face multiple images as input, recent works \cite{arnab2021vivit, bertasius2021space, liu2021decoupled} have produced good performance using ViT as a spatially-decoupled feature extractor and additionally establish a temporally-decoupled feature extractor. They benefit from the fact that video frames are temporally coherent (as shown in Figure~\ref{highlight_video}). These approaches, however, cannot be directly transferred to our task since multi-view reconstruction should deal with unstructured multiple images without prior positional correspondence.
	
	To address this problem, we propose a novel inter-view-decoupled block (IVDB) based on mining the correlation between similar patches from different views (as shown in Figure~\ref{highlight_object}). It can be inserted between the blocks of ViT to create a transformer encoder for unstructured multiple images. This model maintains the advantages of ViT initialization pre-trained on large-scale datasets while alternating decoupling the intra- and inter-view encoding processes. Moreover, by clustering the tokens according to their similarities and exploiting a down-sampling transformer block, the tokens from all branches are compressed into a fixed-size compact representation, ensuring relatively steady performance for the varying number of views input.
	
	In detail, our contributions are as follows:
	
	\begin{itemize}
		\item{To our best knowledge, we are the first to propose a transformer network that alternates decoupled intra- and inter-view feature extraction for multi-view 3D reconstruction, a problem facing unstructured multiple images as input.} 
		\item{Leveraging the correlations between similar tokens, we proposes a novel inter-view-decoupled block (IVDB) that rectifies the tokens according to the related information from other views and a similar-token merger (STM) that compresses the features from all branches.}
		\item{Experiments on ShapeNet \cite{chang2015shapenet} verify that our method achieves performance better than previous SOTA methods by a large margin and has the potential to be more robust for multi-view reconstruction when increasing training consumption.}
	\end{itemize}

	\section{Related Works}
	\label{sec:related_works}
	\subsection{Multi-View 3D Reconstruction}
	
	In the early years, the traditional algorithms, e.g. SFM \cite{ozyecsil2017survey} and SLAM \cite{fuentes2015visual}, build mappings from 2D pixels to 3D positions based on feature matching. They are hard to deal with complex situation of view images. At present, neural network algorithms become the mainstream for solving multi-view reconstruction. 
	
	Among them, CNN-based methods usually extract features from each view in parallel and then aggregate these features to a representation of the shape. Most of the research works focus on the fusion approach. \cite{su2015multi, paschalidou2018raynet, huang2018deepmvs} employ pooling-based fusion method that concatenates the feature maps from different views and then compresses them to a specified size by a maximum pooling or an average pooling layer. Despite being straightforward, it performs poorly because it lacks learnable parameters. 3D-R2N2 series \cite{choy20163d, ma2020improved} and LSM \cite{kar2017learning} use recurrent neural network (RNN)-based fusion method that treats features from views as a sequence. However, a recurrent unit cannot satisfy invariant to permutations and is not suitable for facing a large number of views input due to limited long-term memory. Attsets\cite{yang2020robust}, Pix2Vox series \cite{xie2019pix2vox, xie2020pix2vox++} and GARNet \cite{zhu2023garnet} exploits attention-based fusion method that accumulates the features from different views weighted according to the score maps predicted by an extra branch.
	
	To learn the relatively complex latent correlation between different views, transformer-based methods are proposed. EVolT \cite{wang2021multi}, LegoFormer \cite{yagubbayli2021legoformer} and 3D-C2FT \cite{tiong20223d} treats the input as a sequence on the inter-view-dimension. However, their reconstruction quality is terrible when facing few view images due to the insufficient size of the feature. 3D-RETR \cite{shi20213d} deals with tokens on intra-view-dimension. It utilizes ViT \cite{dosovitskiy2021image} to extract features from each view and fuses them using the pooling-based fusion. Although the strong representation learning ability of the transformer for images is utilized, the potential information between views is not fully discovered.

	\begin{figure*}[ht]
		\centering
		\scalebox{1.2}{
			\includegraphics[width=0.8\linewidth]{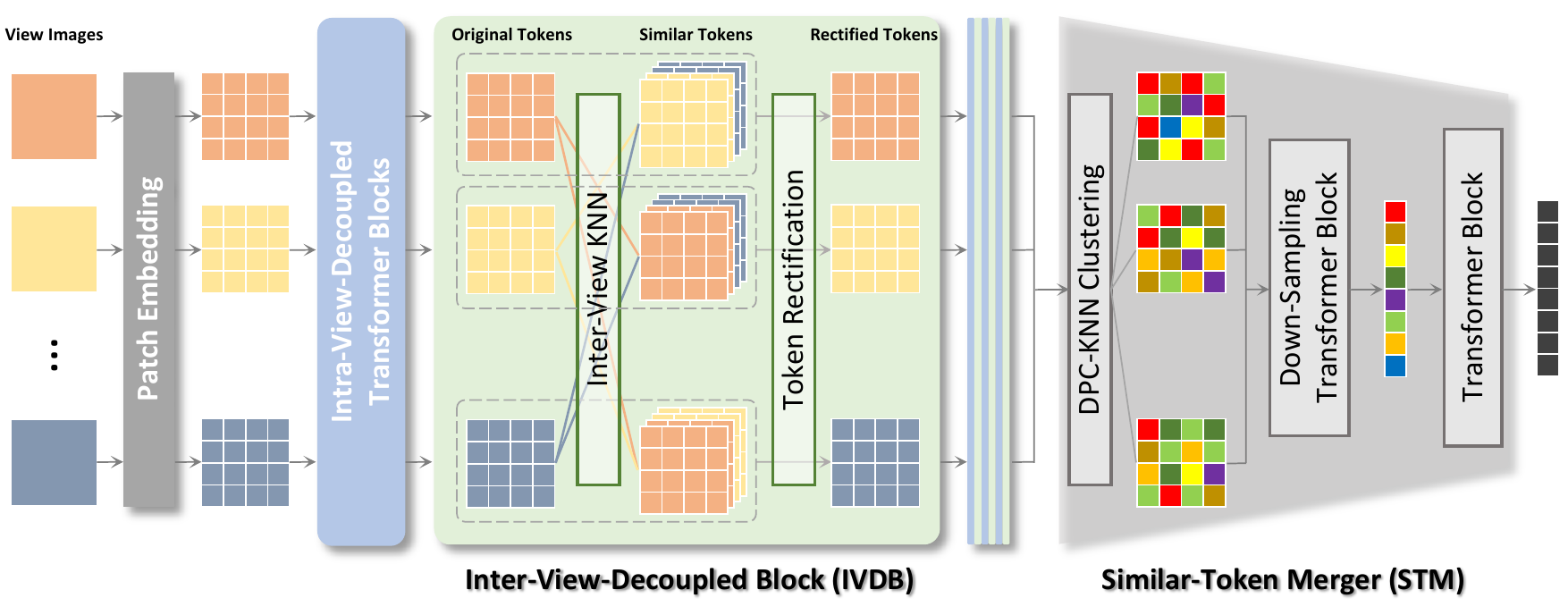}}
		\caption{The architecture used for feature extraction in UMIFormer. The network encodes unstructured multiple images utilizing the intra-view-decoupled transformer block and the inter-view-decoupled block alternately. Then, the feature is compressed to a compact representation by the similar-token merger.}
		\label{pipeline}
	\end{figure*}

	\subsection{Transformer for Multiple Images}
	
	The transformer paradigm is proposed by \cite{vaswani2017attention} for natural language processing. ViT \cite{dosovitskiy2021image} widely extends it to computer vision and mainly works on single-image. Some research about video tasks introduces the transformer network to solve the problems facing multiple images as input. Leveraging prior relationships based on spatial and temporal, TimeSformer~\cite{bertasius2021space} proposes decoupled spatial-temporal attention where the two kinds of attention operation coexist in a transformer block and DSTT~\cite{liu2021decoupled} decouples the spatial and temporal encoding into separate transformer blocks which are used alternately. ViViT~\cite{arnab2021vivit} factorises the multi-head dot-product attention operation to execute the two kinds of decoupled attention in parallel. In addition, DeViT~\cite{cai2022devit}, FGT~\cite{zhang2022flow}, E$^2$FGVI~\cite{li2022towards} and MotionRGBD~\cite{Zhou_2022_CVPR} further mine the relationship in spatial and temporal to acquire a better representation.
	
	These decoupling methods exploit spatial-temporal relations to achieve good representation learning capability. However, they cannot be transferred to solve multi-view reconstruction because there is no inter-image-coherent when processing unstructured multiple images.

	\section{Methods}
	\label{sec:methods}
	
	According to an arbitrary number of view images $\mathcal{I} = \{I_1, I_2, \cdots, I_n\}$ with sizes of $224^2 \times 3$ of an object, our model is to generate the corresponding voxel representation $V$ with a size of $32^3$. To begin, views are used to extract a feature representation $f$ (described in section~\ref{sec:method:feature_extraction}). Then, the binary voxel $V$ will be constructed according to the feature (described in section~\ref{sec:method:shape_reconstruction}). The entire process is formulated as:
	\begin{equation}
		V = \mathtt{UMIFormer}(\mathcal{I}) = \mathtt{R}\left(\mathtt{E}\left(I_1, I_2, \ldots, I_n\right)\right),
	\end{equation}
	where $\mathtt{E}$ and $\mathtt{R}$ denote the processes of feature extraction and shape reconstruction respectively.
	
	\subsection{Feature Extraction}
	\label{sec:method:feature_extraction}
	
	The feature extractor is designed based on ViT \cite{dosovitskiy2021image}. It contains four types of blocks: patch embedding, intra-view-decoupled transformer block, inter-view-decoupled block (IVDB) and similar-token merger (STM). The first two of them are derived from the ViT structure. Patch embedding is consist of splitting images into patches, linearly mapping them and adding position embeddings. The foundation structure of ViT is actually assembled by all the intra-view-decoupled transformer blocks.
	
	IVDBs (elaborated in Section~\ref{sec:method:feature_extraction:ivdb}), which construct the relationships between various views, are periodically inserted into the ViT backbone. Thus, it is feasible to alternate intra- and inter-view-dimension encoding. Comparing to the approaches that separate the two encoding modes, which extract features from each view and then fuse them, our method mines richer correlations between different views. Comparing to the approaches that couple these two dimensions, our method is equivalent to providing prior knowledge to reduce the complexity of representation learning.
	
	STM (elaborated in Section~\ref{sec:method:feature_extraction:stm}), at the end of the extractor, downsamples the feature obtained from all branches into a compact representation. Note that STM is utilized to compress the feature, as opposed to being employed for aggregating like the merger blocks in previous works \cite{su2015multi, choy20163d, yang2020robust}. Because the connection between various views has been created by IVDBs and the transformer-based decoder can handle sequences with variable lengths, it is not necessary to set a specific fusion function. STM is designed for enabling the extractor to provide fixed-size features for reconstruction when receiving varying numbers of views, ensuring relatively stable performance.
	
	The process of feature extraction is shown in Figure~\ref{pipeline}. The patched view images are alternately decoupled encoding by the transformer blocks and IVDBs and then compressed into a compact representation using STM.
	
	\subsubsection{Inter-View-Decoupled Block}
	\label{sec:method:feature_extraction:ivdb}
	
	Since the unstructured images have no prior positional correspondence, we consider building a substitute property between different views. In fact, a token represents a particular region of the object and similar tokens correspond to related regions. Thus, we hold that the relationship created by similar tokens can be used as the auxiliary condition for bridging views. We adopt an inter-view KNN layer, which takes each token as an anchor $x_i$ and matches it with the nearest $k$ tokens $y_{i}$ from each other view in Euclidean space.
	
	\begin{figure}[t]
		\centering
		\scalebox{1.25}{
			\includegraphics[width=0.8\linewidth]{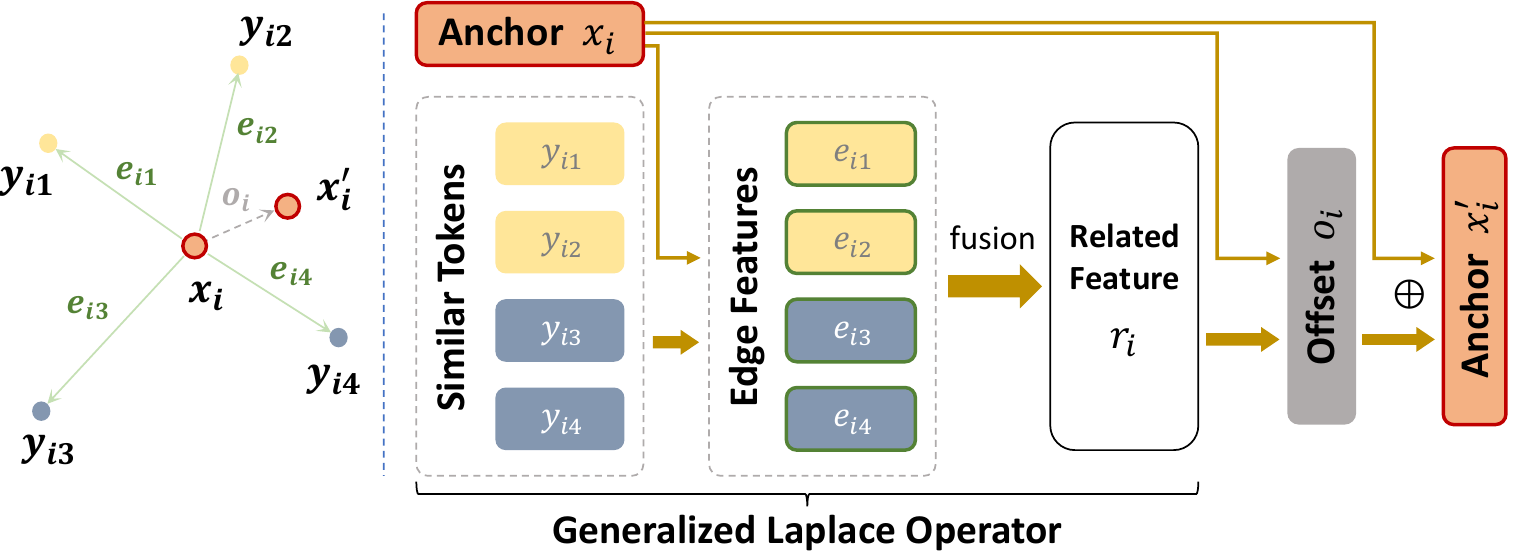}}
		\caption{Visual illustration of the token rectification used in IVDB.}
		\label{token_rectification}
	\end{figure}
	
	To exploit the above relationships, we propose token rectification for mining the inter-view correlation, illustrated in Figure~\ref{token_rectification}. The tokens are treated as point cloud in a high-dimensional manifold space. Among them, the neighboring points, which are regarded as the similar tokens, support the anchor modified to a more accurate representation.
	
	Firstly, a Generalized Laplace Operator extracts the related feature $r_i$ of the anchor $x_i$. This operator embeds the edge features based on the related positional relationship between anchor and its similar tokens in feature space and aggregates them through an attention-based fusion \cite{yang2020robust}. The definition is as follows:
	\begin{equation}
		e_{ij} = \mathtt{MLP_{edge}}\left(y_{ij} - x_{i}\right),
	\end{equation}
	\begin{equation}
		r_{i} = \mathtt{Fusion}\left(e_{i1}, e_{i2}, \ldots, e_{i\left(k\left(n-1\right)\right)}\right),
		\label{equation::fusion}
	\end{equation}
	where $\mathtt{MLP_{edge}}$ is a MLP (multilayer perceptron) with non-linear activation function and $\mathtt{Fusion}$ indicates the attention-based fusion approach. Then, the related feature help to rectify the anchor by predicting the feature offset $o_{i}$. Compared with mapping directly to update the anchor, adding offset better preserves the advantages of ViT initialization pre-trained on large-scale datasets due to keeping the token in the original feature space. We propose two strategies for predicting offsets:
	\begin{equation}
		o_i = \mathtt{MLP_{os}} \left(x_{i}, r_{i}\right),
		\label{offset}
	\end{equation}
	\begin{equation}
		o_i = \mathtt{MLP_{ow}} \left(x_{i}, r_{i}\right) \times x_{i},
		\label{offset_weight}
	\end{equation}
	where $\mathtt{MLP_{os}}$ is an MLP layer to predict the offset directly, while $\mathtt{MLP_{ow}}$ is an MLP with a tanh function to predict the offset weight. Finally, the rectified anchor token is produced by a straightforward addition operation:
	\begin{equation}
		x_{i}^{\prime} = x_{i} + o_{i}.
	\end{equation}
	
	\begin{figure}[t]
		\centering
		\scalebox{1.25}{
			\includegraphics[width=0.78\linewidth]{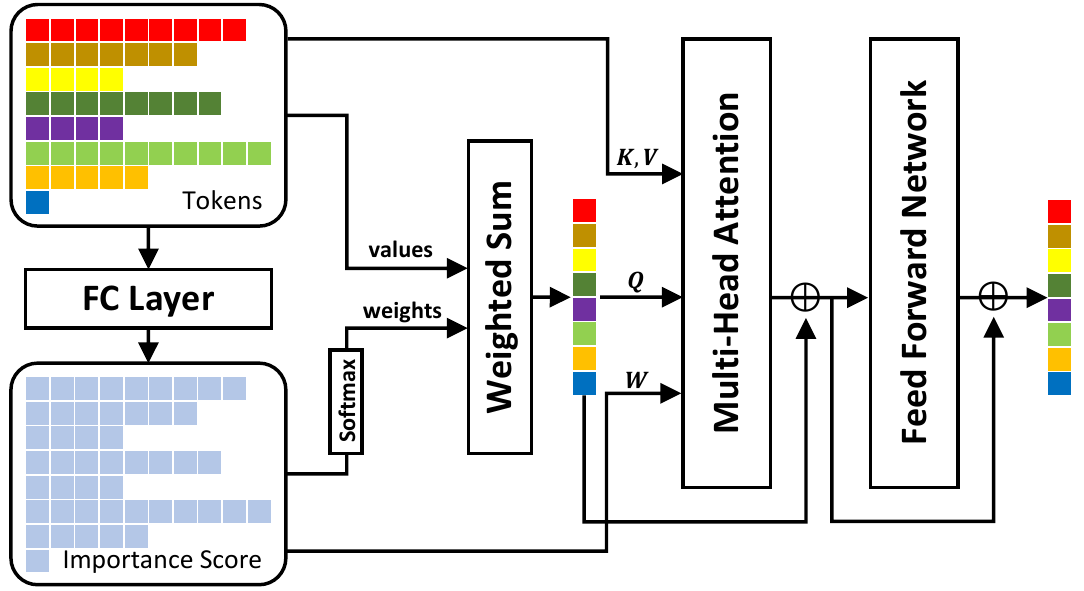}}
		\caption{Visual illustration of the down-sampling transformer block used in STM.}
		\label{down-sampling_transformer_block}
	\end{figure}
	
	\subsubsection{Similar-Token Merger}
	\label{sec:method:feature_extraction:stm}
	
	The previous network fetches a large number of tokens from views. Several of these tokens contain relatively close information that may cause information redundancy, especially when facing extremely high input amounts. Moreover, the view images are composed of both foreground and background while only the foreground information supports reconstruction. As a result, we propose the feature compression method that maximizes diversity while minimizing irrelevant information for the preserved feature.
	
	\begin{table*}[]
		\centering
		\renewcommand\arraystretch{1.5}
		\scalebox{0.63}{
			\begin{tabular}{c|c|ccccccccc}
				\multicolumn{2}{c|}{\textbf{Methods}} & \multicolumn{1}{c}{\textbf{1 view}} & \multicolumn{1}{c}{\textbf{2 views}} & \multicolumn{1}{c}{\textbf{3 views}} & \multicolumn{1}{c}{\textbf{4 views}} & \multicolumn{1}{c}{\textbf{5 views}} & \multicolumn{1}{c}{\textbf{8 views}} & \multicolumn{1}{c}{\textbf{12 views}} & \multicolumn{1}{c}{\textbf{16 views}} & \multicolumn{1}{c}{\textbf{20 views}} \\ \hline
				\multirow{5}{*}{\rotatebox{90}{\textbf{CNN-Based}}} & \textbf{3D-R2N2} \cite{choy20163d} & 0.560 / 0.351 & 0.603 / 0.368 & 0.617 / 0.372 & 0.625 / 0.378 & 0.634 / 0.382 & 0.635 / 0.383 & 0.636 / 0.382 & 0.636 / 0.382 & 0.636 / 0.383 \\
				& \textbf{AttSets} \cite{yang2020robust} & 0.642 / 0.395 & 0.662 / 0.418 & 0.670 / 0.426 & 0.675 / 0.430 & 0.677 / 0.432 & 0.685 / 0.444 & 0.688 / 0.445 & 0.692 / 0.447 & 0.693 / 0.448 \\
				& \textbf{Pix2Vox++} \cite{xie2020pix2vox++} & 0.670 / \textbf{0.436} & 0.695 / 0.452 & 0.704 / 0.455 & 0.708 / 0.457 & 0.711 / 0.458 & 0.715 / 0.459 & 0.717 / 0.460 & 0.718 / 0.461 & 0.719 / 0.462 \\
				& \textbf{GARNet} \cite{zhu2023garnet} & 0.673 / 0.418 & 0.705 / 0.455 & 0.716 / 0.468 & 0.722 / 0.475 & 0.726 / 0.479 & 0.731 / 0.486 & 0.734 / 0.489 & 0.736 / 0.491 & 0.737 / 0.492 \\
				& \textbf{GARNet+} & 0.655 / 0.399 & 0.696 / 0.446 & 0.712 / 0.465 & 0.719 / 0.475 & 0.725 / 0.481 & 0.733 / 0.491 & 0.737 / 0.498 & 0.740 / 0.501 & 0.742 / 0.504 \\ \hline
				\multirow{7}{*}{\rotatebox{90}{\textbf{Transformer-Based}}} & \textbf{EVolT} \cite{wang2021multi} & - / - & - / - & - / - & 0.609 / 0.358 & - / - & 0.698 / 0.448 & 0.720 / 0.475 & 0.729 / 0.486 & 0.735 / 0.492 \\
				& \textbf{Legoformer} \cite{yagubbayli2021legoformer} & 0.519 / 0.282 & 0.644 / 0.392 & 0.679 / 0.428 & 0.694 / 0.444 & 0.703 / 0.453 & 0.713 / 0.464 & 0.717 / 0.470 & 0.719 / 0.472 & 0.721 / 0.472 \\
				& \textbf{3D-C2FT} \cite{tiong20223d} & 0.629 / 0.371 & 0.678 / 0.424 & 0.695 / 0.443 & 0.702 / 0.452 & 0.702 / 0.458 & 0.716 / 0.468 & 0.720 / 0.475 & 0.723 / 0.477 & 0.724 / 0.479 \\
				& \textbf{3D-RETR} \small{(3 views)} & 0.674 / - & 0.707 / - & 0.716 / - & 0.720 / - & 0.723 / - & 0.727 / - & 0.729 / - & 0.730 / - & 0.731 / - \\
				& \textcolor{gray}{\textbf{3D-RETR}\space{$^\dagger$}\space\cite{shi20213d}} & \textcolor{gray}{0.680 / -} & \textcolor{gray}{0.701 / -} & \textcolor{gray}{0.716 / -} & \textcolor{gray}{0.725 / -} & \textcolor{gray}{0.736 / -} & \textcolor{gray}{0.739 / -} & \textcolor{gray}{0.747 / -} & \textcolor{gray}{0.755 / -} & \textcolor{gray}{0.757 / -} \\
				\cline{2-11}
				& \textbf{UMIFormer} & \textbf{0.6802} / 0.4281 & \textbf{0.7384} / \textbf{0.4919} & \textbf{0.7518} / \textbf{0.5067} & 0.7573 / \textbf{0.5127} & 0.7612 / 0.5168 & 0.7661 / 0.5213 & 0.7682 / 0.5232 & 0.7696 / 0.5245 & 0.7702 / 0.5251 \\
				& \textbf{UMIFormer+} & 0.5672 / 0.3177 & 0.7115 / 0.4568 & 0.7447 / 0.4947 & \textbf{0.7588} / 0.5104 & \textbf{0.7681} / \textbf{0.5216} & \textbf{0.7790} / \textbf{0.5348} & \textbf{0.7843} / \textbf{0.5415} & \textbf{0.7873} / \textbf{0.5451} & \textbf{0.7886} / \textbf{0.5466} \\ \hline
		\end{tabular}}
		\caption{Evaluation and comparison of the performance on ShapeNet using IoU $\uparrow$ / F-Score$@1\%$ $\uparrow$. The best results are highlighted in bold. \textcolor{gray}{$^{\dagger}$} The results in this row are derived from models that train individually for the various number of input views.}
		\label{total_result}
	\end{table*}
	
	Inspired by \cite{zeng2022not}, we establish the similarity relationship of tokens again. The tokens are divided into $g$ groups by the DPC-KNN clustering \cite{du2016study} and then fed into the down-sampling transformer block (illustrated in Figure~\ref{down-sampling_transformer_block}). We fuse the features from each group to obtain an aggregated token set using the attention-based fusion method same as $\mathtt{Fusion}$ in Equation~\ref{equation::fusion}. The set is entered into multi-head attention (MHA) as $Q$ to extract information from the ungrouped tokens which provide $K$ and $V$. In contrast to the general MHA, we introduce the extra weights $W$, which reuse the importance score predicted by the additional branch in the attention-based fusion process, to ensure that tokens with different importance have different effects on the result. This MHA is following \cite{zeng2022not} and defined as:
	\begin{equation}
		\mathtt{Attention}(Q, K, V, W)=\mathtt{softmax}\left(\frac{Q K^T}{\sqrt{d_k}}+W\right) V,
	\end{equation}
	where $d_k$ is the dimensions of $Q$, $K$ and $V$. At the end of STM, the down-sampled feature is further processed by a transformer block into the ultimate feature representation extracted from the multiple image views.
	
	It is a great promotion for the compactness of the compressed representation that similar features are stored in one token or a few tokens. In this way, the background tokens with typically limited details are easy to be clustered into the same group with the nearby certain range tokens and then are compressed to a small number of tokens after merging, thereby alleviating the information redundancy.
	
	\subsection{Shape Reconstruction}
	\label{sec:method:shape_reconstruction}
	We employ a decoder composed of a transformer stage and a CNN stage for shape reconstruction, which shares the same structure as \cite{shi20213d}. The transformer stage contains 8 transformer decoder blocks while excluding any upsampling layers. A feature map with a size of $64\times768$ is generated. After reshaping to $4^3\times768$, it entered into the CNN stage and upsample to $32^3$ voxel gradually.
	
	\subsection{Loss Function}
	The task to reconstruct the shape of an object can be seen as a voxel-level segmentation for occupied or empty. Consequently, the loss function is defined as Dice loss \cite{milletari2016v} between predicting volume and the ground truth (GT). The previous work \cite{shi20213d} indicates that it is suitable for 3D reconstruction, the problem with an extremely unbalanced amount of samples between categories. Mathematically, this loss function is defined as:
	\begin{equation}
		\mathcal{L} = 1-\frac{\sum_{i=1}^{32^3}p_{i} gt_{i}}{\sum_{i=1}^{32^3}p_{i}+gt_{i}}-\frac{\sum_{i=1}^{32^3}\left(1-p_{i}\right)\left(1-gt_{i}\right)}{\sum_{i=1}^{32^3}2-p_{i}-gt_{i}}
	\end{equation}
	where $p$ and $gt$ indicate the confidence of the grids on the reconstructed volume and GT.

	\section{Experiments}
	\label{sec:experiments}
	
	\subsection{Datasets and Implementation Details}
	\label{sec:experiments:details}
	
	Following \cite{choy20163d}, our experiments are primarily carried out on a subset of ShapeNet \cite{chang2015shapenet} to evaluate the ability of multi-view 3D reconstruction. The subset includes 13 categories and 43,783 objects with a 3D representation and rendered images from 24 random poses. Moreover, single-view reconstruction experiments on the chair category from Pix3D \cite{sun2018pix3d} dataset including 2,894 data with untruncated and unoccluded view image are supplemented to verify that our model is capable of handling real-world data. The reconstruction results are measured using both 3D Intersection over Union (IoU) and F-Score$@1\%$ \cite{tatarchenko2019single, xie2020pix2vox++}. The evaluation strictly follows the metric used in the related advanced research works to ensure the comparison fairly and reliably.
	
	We adopt the pre-training model of DeiT-B \cite{touvron2021training}, a variant of ViT, to initialize the intra-view-decoupled transformer blocks in our model. To facilitate visualization and analysis, the cls token and distillation token are removed. The model contains 12 transformer blocks and we insert the IVDB with $k=5$ after every third block. For STM, $k_{dpc}$ in DPC-KNN clustering and $g$ are defined as $15$ and $196$. For multi-view 3d reconstruction, we eventually provide two models with the same structure called UMIFormer and UMIFormer+, whose input view numbers are respectively fixed to 3 and 8 during training. The models are trained by an AdamW optimizer \cite{loshchilov2018decoupled} with $\beta_{1}=0.9$ and $\beta_{2}=0.999$ with a batch size of 32 for 150 epochs. The learning rate is defined as 1e-4 and decreases by 0.1 after 50 and 120 epochs sequentially. UMIFormer is trained on 2 Tesla V100 for 2 days and UMIFormer+ is trained on 8 Tesla V100 for 2.5 days. The fixed threshold for binarizing the probabilities is set to 0.5 for UMIFormer and 0.4 for UMIFormer+.

	\begin{figure*}[ht]
		\centering
		\begin{minipage}{0.98\linewidth}
			\centerline{\includegraphics[width=1\linewidth]{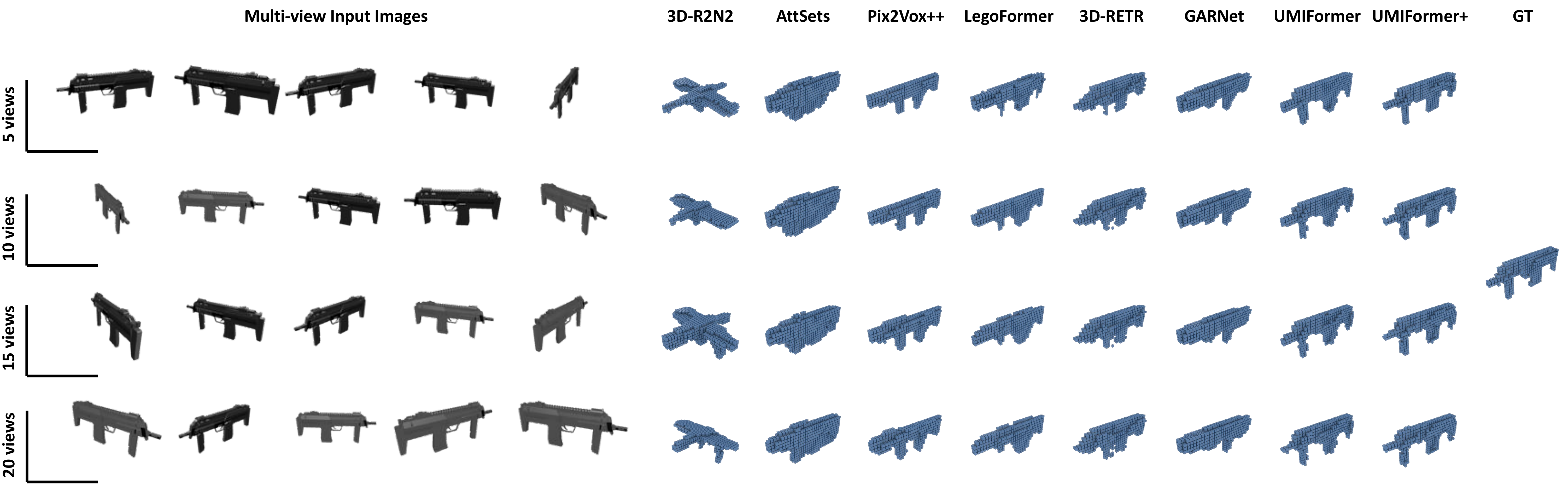}}
		\end{minipage}
		\\ \vspace{0.5mm}
		\begin{minipage}{0.98\linewidth}
			\centerline{\includegraphics[width=1\linewidth]{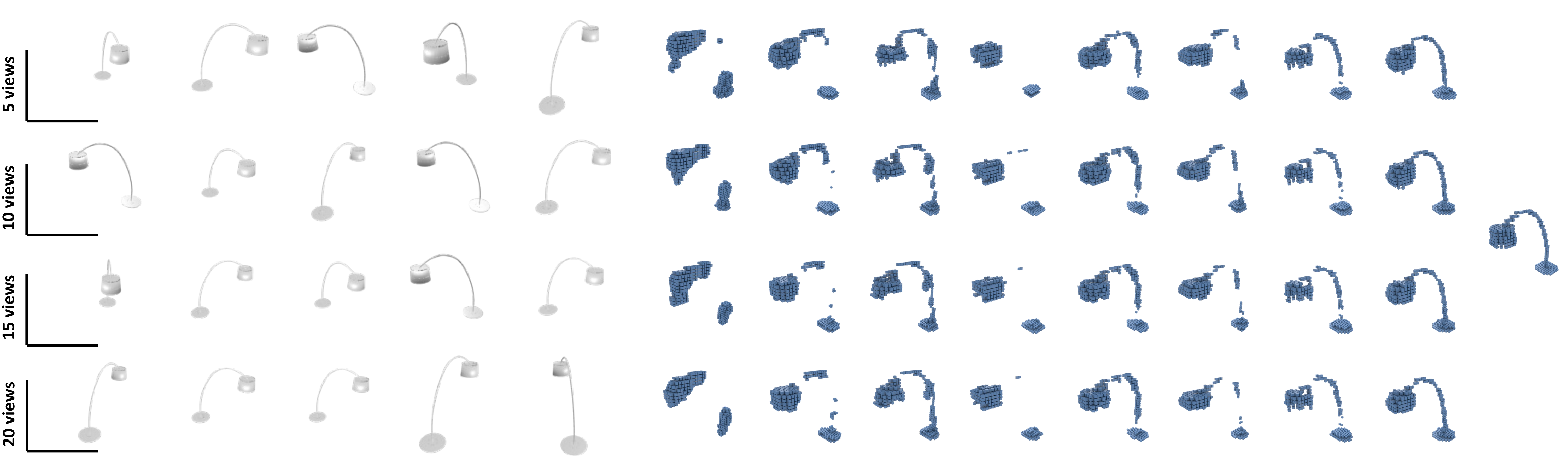}}
		\end{minipage}
		\caption{Multi-view reconstruction results on the test set of ShapeNet when facing 5 views, 10 views, 15 views and 20 views as input. Our method is compared with 3D-R2N2~\cite{choy20163d}, AttSets~\cite{yang2020robust}, Pix2Vox++~\cite{xie2020pix2vox++}, LegoFormer~\cite{yagubbayli2021legoformer}, 3D-RETR~\cite{shi20213d} and GARNet~\cite{zhu2023garnet}.}
		\label{show_results}
	\end{figure*}

	\subsection{Multi-view 3D Reconstruction Results}
	\label{sec:experiments:results}
	
	\begin{table}[]
		\centering
		\renewcommand\arraystretch{1.35}
		\scalebox{0.675}{
			\begin{tabular}{cc|cccccc}
				\multicolumn{1}{c}{\textbf{IVDB}} & \multicolumn{1}{c|}{\textbf{Fusion}} & \multicolumn{1}{c}{\textbf{3 views}} & \multicolumn{1}{c}{\textbf{5 views}} & \multicolumn{1}{c}{\textbf{8 views}} & \multicolumn{1}{c}{\textbf{12 views}} & \multicolumn{1}{c}{\textbf{16 views}} & \multicolumn{1}{c}{\textbf{20 views}} \\ \hline
				\XSolidBrush & \textbf{PBM} & 0.7325 & 0.7406 & 0.7447 & 0.7472 & 0.7487 & 0.7493 \\
				\XSolidBrush & \textbf{ABM} & 0.7394 & 0.7479 & 0.7522 & 0.7545 & 0.7560 & 0.7566 \\
				\XSolidBrush & \textbf{STM} & 0.7477 & 0.7557 & 0.7587 & 0.7598 & 0.7606 & 0.7606 \\
				\Checkmark & \textbf{PBM} & 0.7372 & 0.7453 & 0.7488 & 0.7514 & 0.7530 & 0.7536 \\
				\Checkmark & \textbf{ABM} & 0.7412 & 0.7503 & 0.7548 & 0.7574 & 0.7588 & 0.7593 \\
				\Checkmark & \textbf{STM} & \textbf{0.7518} & \textbf{0.7612} & \textbf{0.7661} & \textbf{0.7682} & \textbf{0.7696} & \textbf{0.7702} \\ \hline
		\end{tabular}}
		\caption{The ablation experiments on ShapeNet evaluated by IoU about IVDB and STM. Among them, STM is compared with two mainstream fusion methods: pooling-based merger (PBM) and attention-based merger (ABM).}
		\label{ablation_experiments}
	\end{table}
	
	The performance qualification results of methods are shown in Table~\ref{total_result}. Undoubtedly, UMIFormer has a significant advantage over the previous methods in almost all metrics. It outperforms current SOTA methods by a large margin. Even training 3D-RETR models separately for different input view numbers (the row marked in gray) trails our model by a big gap. Furthermore, UMIFormer+ boasts a more powerful capability for multi-view reconstruction. Whereas, it has a somewhat limited capacity for single-view reconstruction.
	
	Figure~\ref{show_results} shows two examples of the reconstruction results when facing various view amounts as inputs. Compared with the other methods, our two models produce more accurate results for rifle reconstruction. In addition, the texture on predicted volumes is adjusted slightly and optimized gradually with increasing view inputs. It not only demonstrates the effectiveness of our algorithm but also verifies that our model can continue to mine information from the increasing input. For lamp reconstruction, our models, especially UMIFormer+, realize a relatively complete representation for the intermediate bracket, which is difficult for the other methods. Certainly, it also demonstrates the strong learning ability of our feature extractor for details.

	\subsection{Ablation Experiments}
	
	Ablation analysis on IVDB and STM is based on experimental results as shown in Table~\ref{ablation_experiments}.
	
	\textbf{Effect of IVDB.} We observe that employing IVDB can consistently improve the reconstruction performance for various amounts of view inputs. Table~\ref{rectification_ablation} presents experimental results related to several rectification strategies (discussed in Section~\ref{sec:method:feature_extraction:ivdb}) used in IVDB. Token rectification by mapping directly performs terribly. It maps tokens to a new feature space that does not match the prior knowledge of the backbone network learned during pre-training. As a result, the benefits of pre-training are significantly disrupted. However, the other two strategies adjust tokens on their original feature space to avoid the problem and achieve the expected performance. Among them, predicting offset works better for the case of few input views and predicting offset weight is suitable for processing a large number of input views. In this paper, we adopt predicting offset weight as the default setting.
	
	\textbf{Effect of STM.} All three types of merger — pooling-based merger \cite{su2015multi}, attention-based merger \cite{yang2020robust} and our proposed STM — compress features from all branches to a fixed size of $196\times768$ in our network. Notably, the model using STM can achieve better reconstruction performance. It verifies that STM preserves richer information than other compression methods.
	
	\begin{table}[]
		\centering
		\renewcommand\arraystretch{1.35}
		\scalebox{0.68}{
			\begin{tabular}{c|cccccc}
				\multicolumn{1}{c|}{\textbf{\makecell{Rectification\\Strategy}}} & \multicolumn{1}{c}{\textbf{3 views}} & \multicolumn{1}{c}{\textbf{5 views}} & \multicolumn{1}{c}{\textbf{8 views}} & \multicolumn{1}{c}{\textbf{12 views}} & \multicolumn{1}{c}{\textbf{16 views}} & \multicolumn{1}{c}{\textbf{20 views}} \\ \hline
				\textbf{FC Mapping} & 0.6935 & 0.7022 & 0.7049 & 0.7038 & 0.7028 & 0.7022 \\
				\textbf{Offset} & \textbf{0.7546} & \textbf{0.7632} & \textbf{0.7672} & \textbf{0.7688} & 0.7695 & 0.7694 \\
				\textbf{Offset Weight} & 0.7518 & 0.7612 & 0.7661 & 0.7682 & \textbf{0.7696} & \textbf{0.7702} \\ \hline
		\end{tabular}}
		\caption{Comparison of performance on ShapeNet evaluated by IoU when using different rectification strategies in IVDB. FC mapping refers to using a fully connected layer to map the concatenation of an anchor and its related features to the rectified token directly. Offset and offset weight respectively correspond to the strategies defined in Equation~\ref{offset} and \ref{offset_weight}.}
		\label{rectification_ablation}
	\end{table}
	
	Furthermore, experimental results indicate that using IVDB and STM together performs much better than using them alone.

	\subsection{Evaluation on Real-World Dataset}
	
	Pix3D dataset \cite{sun2018pix3d} is usually used as the testing set for evaluating the performance of single-view reconstruction. Most of the view images in it are real-world images with complex backgrounds. Therefore, we attempt to validate the effectiveness of UMIFormer on this dataset to verify that it works on various domains of image. Following previous works, the training set uses the data from the chair category in ShapeNet and the view images are re-synthesized by Render for CNN \cite{su2015render} with random backgrounds from the SUN database \cite{xiao2010sun}. Among them, each object includes 60 view images.
	
\begin{figure}[ht]
		\centering
		\scalebox{1.25}{
			\includegraphics[width=0.8\linewidth]{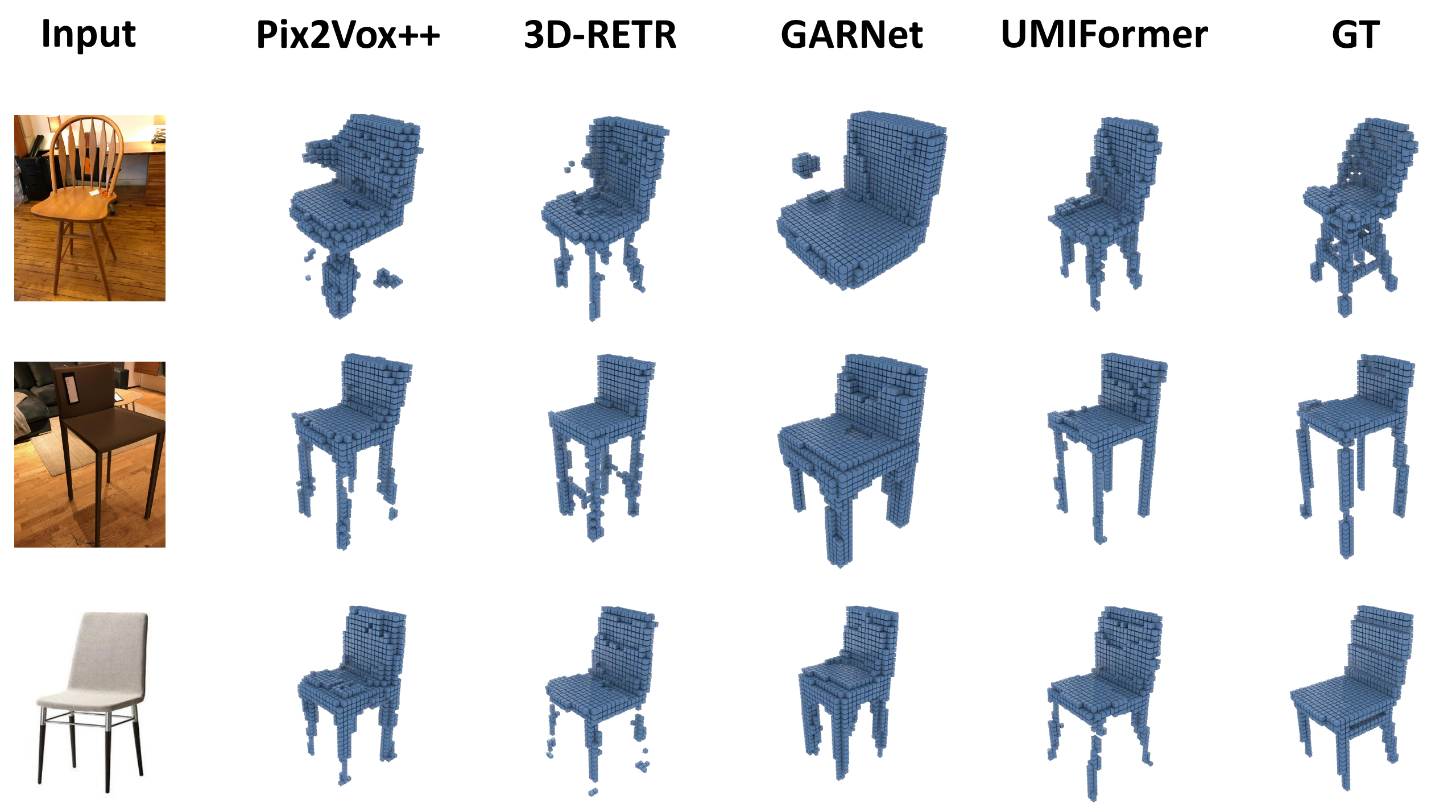}}
		\caption{Single-view reconstruction results on real-world data (Pix3D test set).}
		\label{show_results_pix3d}
	\end{figure}

	IVDB is not used in this experiment because it is irrelevant for single-view input. In addition, $g$ in STM is defined as 32. Table~\ref{pix3d_results} shows the performance qualification results of UMIFormer and other SOTA methods which can be used for both single-view reconstruction and multi-view reconstruction. In comparison, our model slightly outperforms them. Figure~\ref{show_results_pix3d} shows that our method performs better on the restoration of texture, particularly for the thin strip shapes.
	
	\begin{table}[]
		\centering
		\renewcommand\arraystretch{1.}
		\scalebox{0.75}{
			\begin{tabular}{c|c|c|c}
				\hline
				\multicolumn{4}{c}{\textbf{IoU $\uparrow$}} \\ \hline
				\textbf{Pix2Vox++} \cite{xie2020pix2vox++} & \textbf{3D-RETR} \cite{shi20213d} & \textbf{GARNet} \cite{zhu2023garnet} & \textbf{UMIFormer} \\ \hline
				0.279 & 0.297 & 0.291 & \textbf{0.300} \\ \hline \hline
				\multicolumn{4}{c}{\textbf{F-Score$@1\%$ $\uparrow$}} \\ \hline
				\textbf{Pix2Vox++} & \textbf{3D-RETR} & \textbf{GARNet} & \textbf{UMIFormer} \\ \hline
				0.113 & 0.125 & 0.116 & \textbf{0.129} \\ \hline
		\end{tabular}}
		\caption{Evaluation and comparison of the performance for single-view reconstruction on Pix3D.}
		\label{pix3d_results}
	\end{table}
	
	\subsection{Visualization of Similar Tokens}
	
\begin{figure}[t]
		\centering
		\scalebox{1.}{
			\includegraphics[width=0.96\linewidth]{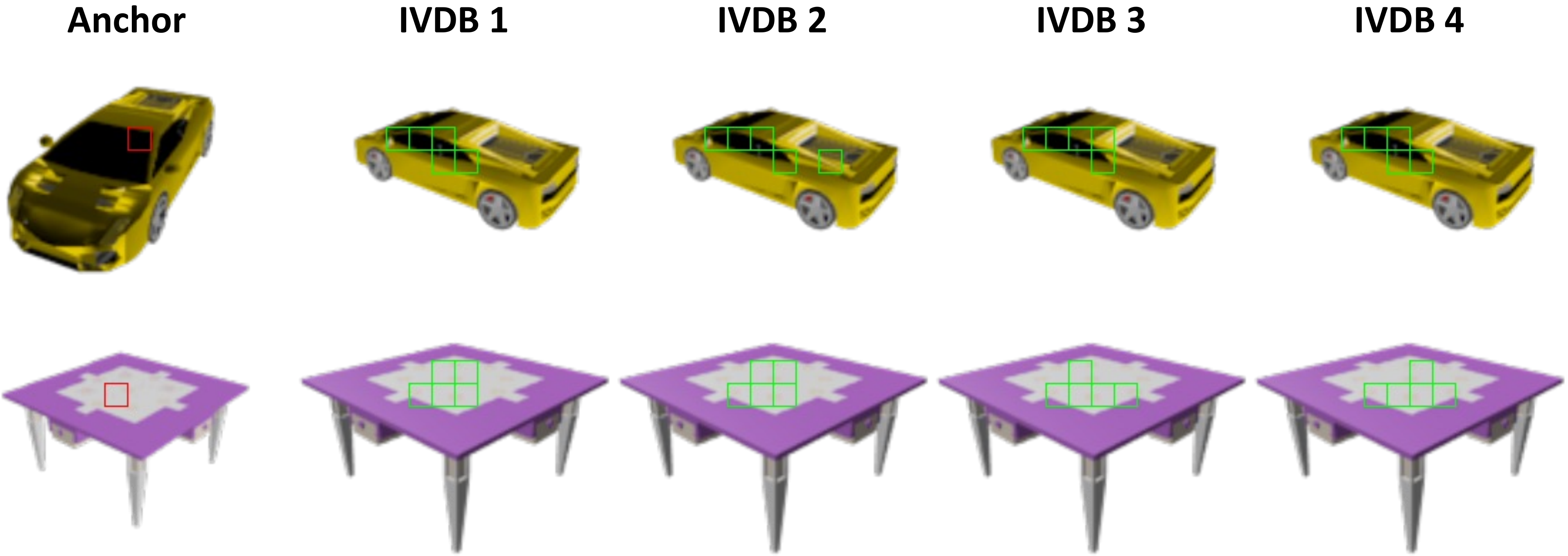}}
		\caption{Visualization of several anchors (marked with red frames) and their similar tokens (marked with green frames) paired by inter-view KNN in IVDBs during multi-view reconstruction processing.}
		\label{visualization_ivdb}
	\end{figure}

	In our algorithm, mining the correlations between similar tokens is involved several times. To further investigate the behavior of our algorithm, we attempt to visualize the pertinent procedures. Taking the UMIFormer dealing with 2 view inputs from ShapeNet as an example, Figure~\ref{visualization_ivdb} shows the results that an anchor (marked with a red frame) finds its similar tokens (marked with green frames) from another view image through the inter-view KNN layers in the 4 IVDBs. We observe that regions relating to the anchor content are surrounded by the matched tokens. Therefore, there is indeed a semantic relationship between them, which can be used as the position correspondence for decoupling.
	
\begin{figure}[ht]
		\centering
		\scalebox{1.25}{
			\includegraphics[width=0.8\linewidth]{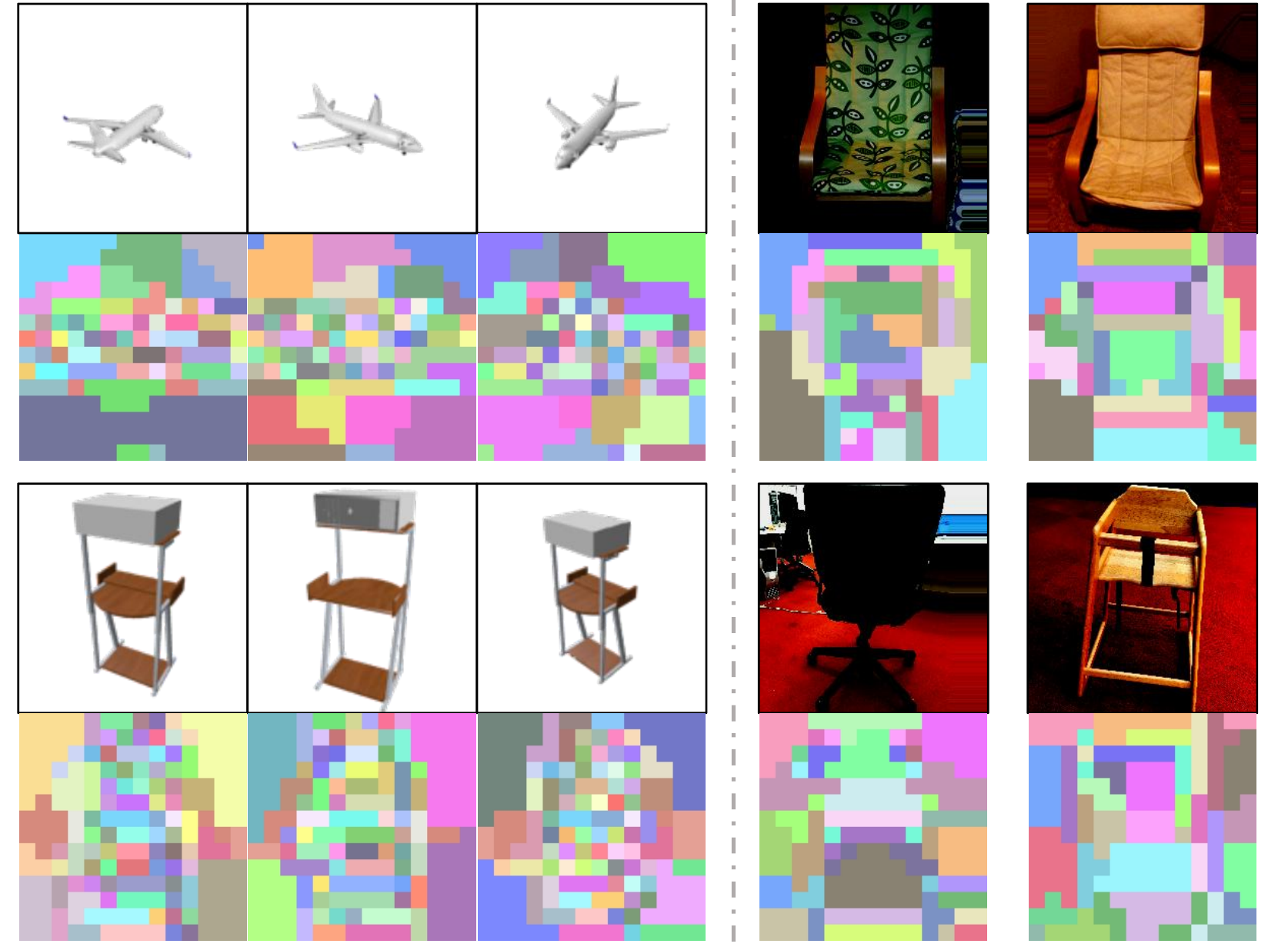}}
		\caption{Visualization of clustering results in STM, including the case of multi-view reconstruction on ShapeNet (left) and single-view reconstruction on Pix3D (right).}
		\label{visualization_stm}
	\end{figure}

	Figure~\ref{visualization_stm} shows some examples of token grouping by the clustering layer in STM. As anticipated, the patches corresponding to the foreground regions are divided finely, whereas the background patches are only assigned to a few groups. Nevertheless, cluster maps do not clearly distinguish subject edges since the features for clustering have been highly abstracted by the previous layers. It does not affect STM to maximize diversity while minimizing irrelevant information at the feature level.
	
	\begin{table*}[]
		\centering
		\renewcommand\arraystretch{1.1}
		\scalebox{0.75}{
			\begin{tabular}{c|c|ccccccccc}
				\multicolumn{2}{c|}{\textbf{Encoder}} & \multicolumn{1}{c}{\textbf{1 view}} & \multicolumn{1}{c}{\textbf{2 views}} & \multicolumn{1}{c}{\textbf{3 views}} & \multicolumn{1}{c}{\textbf{4 views}} & \multicolumn{1}{c}{\textbf{5 views}} & \multicolumn{1}{c}{\textbf{8 views}} & \multicolumn{1}{c}{\textbf{12 views}} & \multicolumn{1}{c}{\textbf{16 views}} & \multicolumn{1}{c}{\textbf{20 views}}\\ \hline
				\multicolumn{2}{c|}{\textbf{Independent Branches}} & \textbf{0.6923} & 0.7292 & 0.7394 & 0.7445 & 0.7479 & 0.7522 & 0.7545 & 0.7560 & 0.7566 \\ \hline
				\multirow{4}{*}{{\textbf{\makecell{Video\\ Transformer}}}} & {\textbf{Joint Attention} \cite{zeng2020learning}} & 0.6771 & 0.7112 & 0.7201 & 0.7241 & 0.7264 & 0.7275 & 0.7269 & 0.7261 & 0.7252 \\
				& {\textbf{Factorised Transformer Block} \cite{liu2021decoupled}} & 0.6809 & 0.7179 & 0.7288 & 0.7337 & 0.7365 & 0.7403 & 0.7424 & 0.7437 & 0.7441 \\
				& \textbf{Factorised Attention} \cite{bertasius2021space} & 0.6918 & 0.7287 & 0.7400 & 0.7451 & 0.7485 & 0.7528 & 0.7552 & 0.7566 & 0.7572 \\
				& \textbf{Factorised Dot-Product} \cite{arnab2021vivit} & 0.6684 & 0.7139 & 0.7275 & 0.7331 & 0.7373 & 0.7423 & 0.7445 & 0.7457 & 0.7463 \\ \hline
				\multicolumn{2}{c|}{\textbf{Ours}} & 0.6802 & \textbf{0.7384} & \textbf{0.7518} & \textbf{0.7573} & \textbf{0.7612} & \textbf{0.7661} & \textbf{0.7682} & \textbf{0.7696} & \textbf{0.7702} \\ \hline
		\end{tabular}}
		\caption{Comparison of the reconstruction performance exploiting different decoupling strategies in feature extractor. To control the variables, all of them are based on the structure of ViT. For more setup details refer to the supplementary material.}
		\label{multi-view_input}
	\end{table*}
	
	\section{Discussion}
	\label{sec:discussion}
	
	In this work, we expect to leverage ViT and decoupled encoding strategy to establish a robust representation learning model for multiple image inputs. The effectiveness of the model has been verified by several works \cite{arnab2021vivit, bertasius2021space, liu2021decoupled} about video transformer algorithms. Although both video tasks and multi-view reconstruction are oriented to multi-image input problems, their inter-image relationships are quite different. The inter-view-decoupled encoder established in this way is theoretically not suitable for processing unstructured multiple images.
	
	In Table~\ref{multi-view_input}, we compare the performance of multi-view 3D reconstruction using different feature extraction methods. Encoding for independent branches using ViT as shown in the first row treats as the baseline. To control the variables, the four video transformer methods are all implemented based on ViT, which may differ from the architecture in the original paper. Encoder with joint attention means that all tokens from various views are processed uniformly without decoupled encoding. We observe that the performance is significantly worse than the baseline and the reconstruction results are worse when dealing with more than 8 views as input. Since the positional encoding for each image in a multi-view reconstruction is consistent, the prior association between tokens in the attention layer is completely lost. It is very challenging to mine the correlation solely relying on the adaptive capability of the network. This problem becomes more serious as the number of views increases. The other three video transformer approaches respectively use factorised transformer block, factorised attention layers and factorised dot products in the self-attention layer to establish decoupled encoding. These methods increase the connection between branches relative to the baseline while they do not conform to the nature of unstructured multiple images. Actually, these kinds of inter-view-decouple feature extraction methods do not simplify the multi-view reconstruction problem. Therefore, these methods cannot achieve the desired performance. Summarizing these results, we confirm that the inter-view-decoupled strategy based on mining the correlation between similar tokens is most suitable for handling unstructured multiple images.

	\section{Conclusion}
	
	\textbf{Limitations.} 1) Our model requires large memory occupation since it is based on the parallel transformer network and devoted to a 3D reconstruction task. Therefore, it is hard to generalize to high-resolution voxel reconstruction under existing mainstream hardware devices. 2) The computational consumption of both the inter-view KNN layer and the DPC-KNN clustering layer grows exponentially with the increasing number of input views. Therefore, the predicting efficiency of our algorithm is not dominant when facing an extremely high number of view inputs.
	
	In this paper, we propose a transformer-based method for multi-view 3D reconstruction which achieves brilliant performance. The feature extractors alternate decoupled intra- and inter-view encoding for unstructured multiple images by mining the correlation between similar tokens. In future work, we expect to overcome its limitation on higher-resolution reconstruction by compressing the model and alleviate the inference efficiency problem by accelerating KNN and DPC-KNN clustering algorithm. Furthermore, this encoding mode may also be extended to other issues involving unstructured multiple images.

	\section*{Acknowledgements}
	
	\setlength{\parindent}{0pt} {This work was supported by National Key Research and Development Plan under Grant 2021YFE0205700, Science and Technology Development Fund of Macau (0070/2020/AMJ, 0004/2020/A1) and Guangdong Provincial Key R\&D Programme: 2019B010148001.}
	
	{\small
		\bibliographystyle{ieee_fullname}
		\bibliography{3d_reconstruction}
	}
	
	\newpage
	\onecolumn
	
	\begin{center}
		\huge{Supplementary Material}
	\end{center}
	
	\setcounter{section}{0}
	\section{Voxel Representation}
	In this paper, our method employs voxel-based 3D representation. It is not necessary to use any extra information for reconstruction. However, methods using implicit representation require camera parameters and methods with mesh representation need an assistive tool.

	\section{Evaluation Metrics}
	In Section~\ref{sec:experiments:details}, we mention that Intersection over Union (IoU) and F-Score$@1\%$ are used as the evaluation metrics to measure the performance of methods. It is emphasized that these two metrics are commonly used and recognized in works related to voxel-based 3D reconstruction. Their details are also kept the same as previous research works and will be elaborated as follow:
	
	\textbf{Intersection over Union.} The predicted probabilities $p$ should be binarized according to a preset threshold and then compare the voxel grids with the ground truth $gt$, which is defined as:
	\begin{equation}
		\text{IoU}=\frac{\sum_{(i,j,k)} \text{I}(p_{(i,j,k)}>t)\text{I}(gt_{(i,j,k)})}{\sum_{i,j,k}\text{I}[\text{I}(p_{(i,j,k)}>t)+\text{I}(gt_{(i,j,k)})]},
	\end{equation}
	where $\text{I}(\cdot)$ is an indicator function and the subscript $\left(i, j, k\right)$ denotes the occupancy probability of the grid located on the corresponding position.
	
	\textbf{F-Score$@1\%$.} \cite{tatarchenko2019single} firstly proposes F-Score and \cite{xie2020pix2vox++} introduces it as an extra metric to 3D reconstruction task. F-Score is defined as:
	\begin{equation}
		\text { F-Score }(d)=\frac{2 \mathrm{P}(d) \mathrm{R}(d)}{\mathrm{P}(d)+\mathrm{R}(d)},
	\end{equation}
	where $P\left(d\right)$ and $R\left(d\right)$ indicate the precision and recall while the distance threshold is $d$. The precision and recall can be calculated as:
	\begin{equation}
		\mathrm{P}(d)=\frac{1}{|\mathcal{R}|} \sum_{\mathbf{r} \in \mathcal{R}}\left[{min} _{\mathbf{g} \in \mathcal{G}}\|\mathbf{r}-\mathbf{g}\|<d\right],
	\end{equation}
	\begin{equation}
		\mathrm{R}(d)=\frac{1}{|\mathcal{R}|} \sum_{\mathbf{r} \in \mathcal{G}}\left[{min} _{\mathbf{r} \in \mathcal{R}}\|\mathbf{g}-\mathbf{r}\|<d\right],
	\end{equation}
	where $\mathcal{R}$ and $\mathcal{G}$ are respectively the point clouds of the reconstruction object and ground truth. The surface of voxel objects is generated by the marching cubes algorithm \cite{lorensen1987marching}. The point clouds with 8192 points sampled from the surface are utilized to calculate F-Score. F-Score$@1\%$ means the F-Score value when $d=1\%$. The above settings are completely consistent with \cite{xie2020pix2vox++}.

	\section{Setup Details of Experiments in Table~\ref{multi-view_input}}
	The experiments as shown in Table~\ref{multi-view_input} attempt to verify that the inter-view-decoupled strategy based on mining the correlations between similar tokens is most suitable for handling unordered multiple images in the multi-view 3D reconstruction task. Consequently, the control groups adapt different encoding strategies and end up with attention-based fusion \cite{yang2020robust}, the most advanced aggregate approach before our work, to connect to the shape reconstruction stage. In addition, all methods are based on ViT \cite{dosovitskiy2021image} with 12 transformer blocks to manipulate the variables. The setup details of them are as follows:
	
	\begin{itemize}
		\item\textbf{Independent Branches.} As the baseline, ViT \cite{dosovitskiy2021image} extracts the feature from view images in parallel while there is no communication between each branch until the fusion module.
		
		\item\textbf{Video Transformer with Joint Attention} \cite{zeng2020learning}\textbf{.} All tokens from various views are processed uniformly in each attention layers without decoupled encoding.
		
		\item\textbf{Video Transformer.} The three methods establish intra-view-decoupled encoding utilizing ViT and insert inter-view-decoupled encoding based on the temporally-coherence property in different ways.
		\begin{itemize}
			\item\textbf{Factorised Transformer Block} \cite{liu2021decoupled}\textbf{.} The 12 transformer blocks are divided into 2 types: intra-view-decoupled transformer blocks and inter-view-decoupled transformer blocks. The former independently processes each image. Whereas, the latter processes on different images at the same spatial regions and the regions are defined to a size of $7 \times 7$ tokens without overlap. These two types of blocks are executed alternately.
			
			\item\textbf{Factorised Attention} \cite{bertasius2021space}\textbf{.} An extra multi-head attention (MHA) layer initialised with zero for its all weights is inserted between the original MHA layer and feed-forward network in each transformer block. The original MHA layer computes self-attention on the intra-view-dimension and the inserted MHA layer does on the inter-view-dimension.
			
			\item\textbf{Factorised Dot-Product} \cite{arnab2021vivit}\textbf{.} In each MHA, intra-view-decoupled attention operation and inter-view-decoupled attention operation are factorised to compute using different heads in parallel. This method has the same number of parameters as ViT.
		\end{itemize}
		\item\textbf{Ours.} This experiment adapts the UMIFormer model.
	\end{itemize}
	
	Obviously, the inter-view-decoupled strategies used in video transformer networks do not conform to the nature of unordered multiple images, while the strategy based on mining the correlations between similar tokens can establish a relatively reasonable connection between views. Therefore, our proposed inter-view-decoupled strategy is more suitable for multi-view 3D reconstruction, as demonstrated by the experiment results.

	\section{Number of View Input during Training}
	
	For multi-view reconstruction algorithms, the performance to process a heavy amount of input can be better by increasing the view number during training. All SOTA methods we compared in Table~\ref{total_result} employ not less than 3 views as input for training. Among them, the training view number of AttSets~\cite{yang2020robust} even attain 24. Therefore, the effectiveness of UMIFormer can be fully verified through the great multi-view reconstruction performance when only adopting 3 views during training.

	\section{Supplementary Experiments}
	
	\subsection{24-View Reconstruction Results}
	
	As a work on multi-view 3D reconstruction, it is necessary to pay attention to the performance when facing a large number of view inputs. As shown in Table~\ref{24-view_results}, we take 24 view inputs as an example. UMIFormer and UMIFormer+ have significant advantages over the other SOTA methods in terms of reconstructing each category.
	
	\begin{table*}[b]
		\centering
		\renewcommand\arraystretch{1.43}
		\scalebox{0.625}{
			\begin{tabular}{c|cccccc|cccccc}
				& \multicolumn{6}{c|}{\textbf{24-view IoU}} & \multicolumn{6}{c}{\textbf{24-view F-Score@1\%}} \\ \hline
				\textbf{Category} & \textbf{Pix2Vox++ \cite{xie2020pix2vox++}} & \textbf{EVolT\cite{wang2021multi}} & \textbf{GARNet} \cite{zhu2023garnet} & \textbf{GARNet+} & \textbf{UMIFormer} & \textbf{UMIFormer+} & \textbf{Pix2Vox++} & \textbf{EVolT} & \textbf{GARNet} & \textbf{GARNet+} & \textbf{UMIFormer} & \textbf{UMIFormer+} \\ \hline
				airplane & 0.729 & 0.741 & 0.724 & 0.739 & 0.769 & \textbf{0.789} & 0.614 & 0.636 & 0.606 & 0.628 & 0.667 & \textbf{0.691} \\
				bench & 0.686 & 0.707 & 0.698 & 0.707 & 0.738 & \textbf{0.761} & 0.522 & 0.548 & 0.536 & 0.551 & 0.498 & \textbf{0.600} \\
				cabinet & 0.829 & 0.832 & 0.841 & 0.840 & 0.861 & \textbf{0.877} & 0.456 & 0.464 & 0.473 & 0.473 & 0.498 & \textbf{0.515} \\
				car & 0.883 & 0.894 & 0.888 & 0.894 & 0.895 & \textbf{0.903} & 0.598 & 0.624 & 0.608 & 0.623 & 0.622 & \textbf{0.641} \\
				chair & 0.647 & 0.681 & 0.674 & 0.683 & 0.713 & \textbf{0.735} & 0.341 & 0.373 & 0.369 & 0.384 & 0.399 & \textbf{0.419} \\
				display & 0.613 & 0.674 & 0.668 & 0.665 & 0.742 & \textbf{0.768} & 0.335 & 0.403 & 0.386 & 0.396 & 0.454 & \textbf{0.485} \\
				lamp & 0.493 & 0.520 & 0.516 & 0.513 & 0.570 & \textbf{0.610} & 0.351 & 0.366 & 0.366 & 0.369 & 0.410 & \textbf{0.451} \\
				speaker & 0.762 & 0.772 & 0.773 & 0.772 & 0.820 & \textbf{0.840} & 0.326 & 0.339 & 0.338 & 0.346 & 0.392 & \textbf{0.418} \\
				rifle & 0.686 & 0.711 & 0.697 & 0.709 & 0.760 & \textbf{0.784} & 0.624 & 0.653 & 0.634 & 0.647 & 0.707 & \textbf{0.736} \\
				sofa & 0.782 & 0.800 & 0.807 & 0.810 & 0.825 & \textbf{0.840} & 0.454 & 0.478 & 0.489 & 0.500 & 0.505 & \textbf{0.528} \\
				table & 0.666 & 0.675 & 0.693 & 0.692 & 0.726 & \textbf{0.744} & 0.419 & 0.431 & 0.449 & 0.452 & 0.467 & \textbf{0.481} \\
				telephone & 0.849 & 0.867 & 0.871 & 0.879 & 0.887 & \textbf{0.904} & 0.666 & 0.687 & 0.698 & 0.716 & 0.709 & \textbf{0.736} \\
				watercraft & 0.668 & 0.693 & 0.693 & 0.696 & 0.723 & \textbf{0.745} & 0.460 & 0.494 & 0.494 & 0.504 & 0.534 & \textbf{0.567} \\ \hline
				\textbf{Overall} & 0.720 & 0.738 & 0.737 & 0.742 & 0.771 & \textbf{0.790} & 0.473 & 0.497  & 0.493 & 0.505 & 0.525 & \textbf{0.548} \\ \hline
		\end{tabular}}
		\caption{Evaluation and comparison of the performance for 24-view reconstruction on the test set of ShapeNet using IoU / F-Score@1\%.}
		\label{24-view_results}
	\end{table*}
	
	\subsection{More Reconstruction Examples}
	
\begin{figure*}[ht]
		\centering
		\begin{minipage}{1.03\linewidth}
			\centerline{\includegraphics[width=1\linewidth]{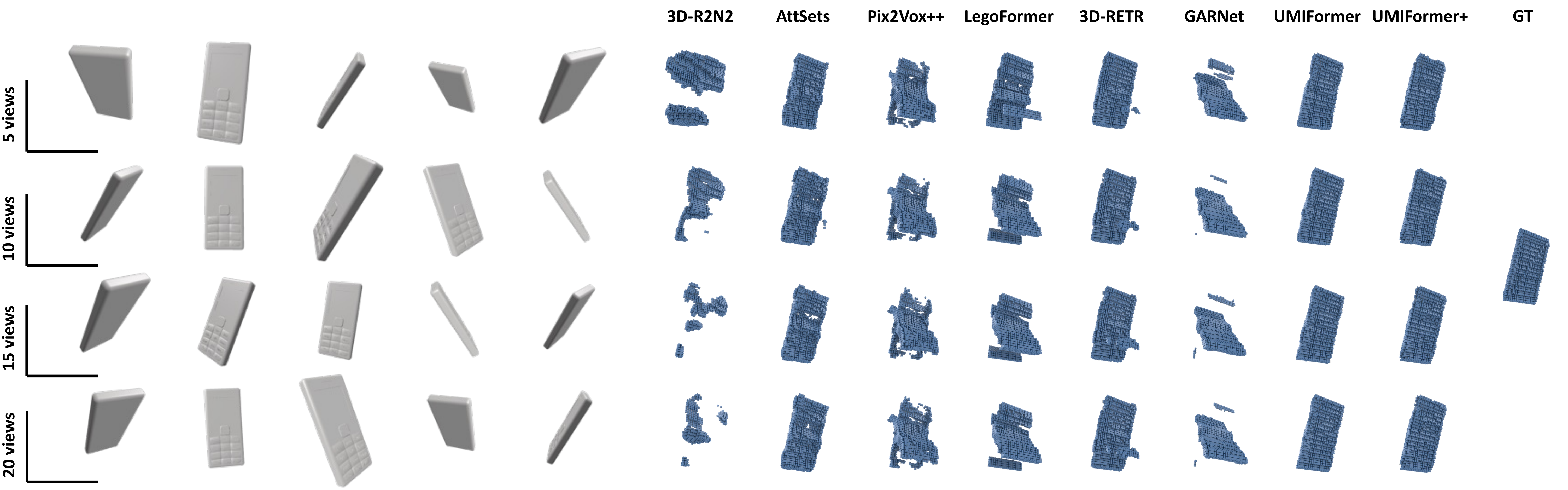}}\hspace{1000mm}
		\end{minipage}
		\\
		\begin{minipage}{1.03\linewidth}
			\centerline{\includegraphics[width=1\linewidth]{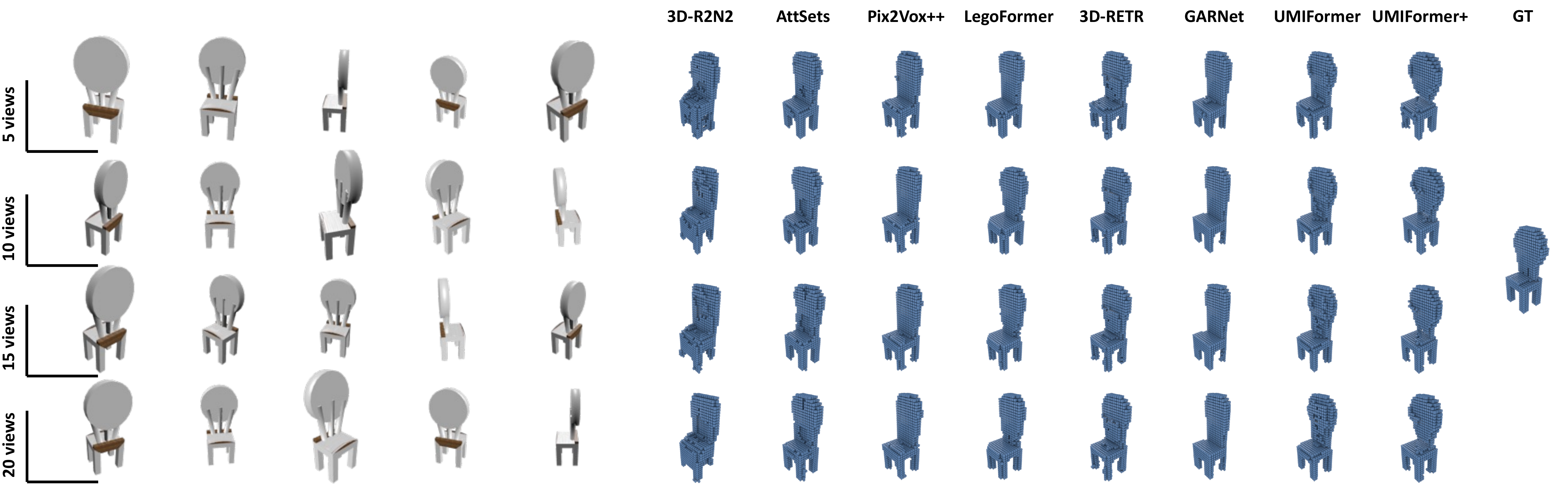}}\hspace{1000mm}
		\end{minipage}
		\\
		\begin{minipage}{1.03\linewidth}
			\centerline{\includegraphics[width=1\linewidth]{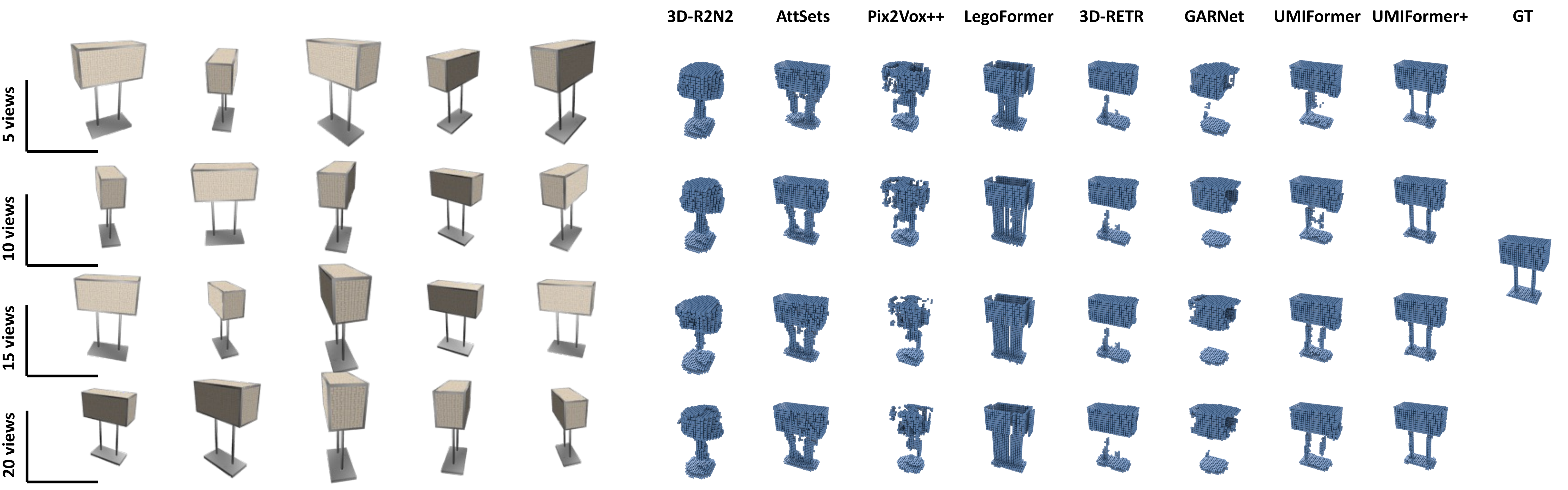}}
		\end{minipage}
		\caption{Qualitative reconstruction results when facing 5 views, 10 views, 15 views and 20 views as input for telephone, chair and lamp.}
		\label{show_results_1}
	\end{figure*}

\begin{figure*}[ht]
		\centering
		\begin{minipage}{1.03\linewidth}
			\centerline{\includegraphics[width=1\linewidth]{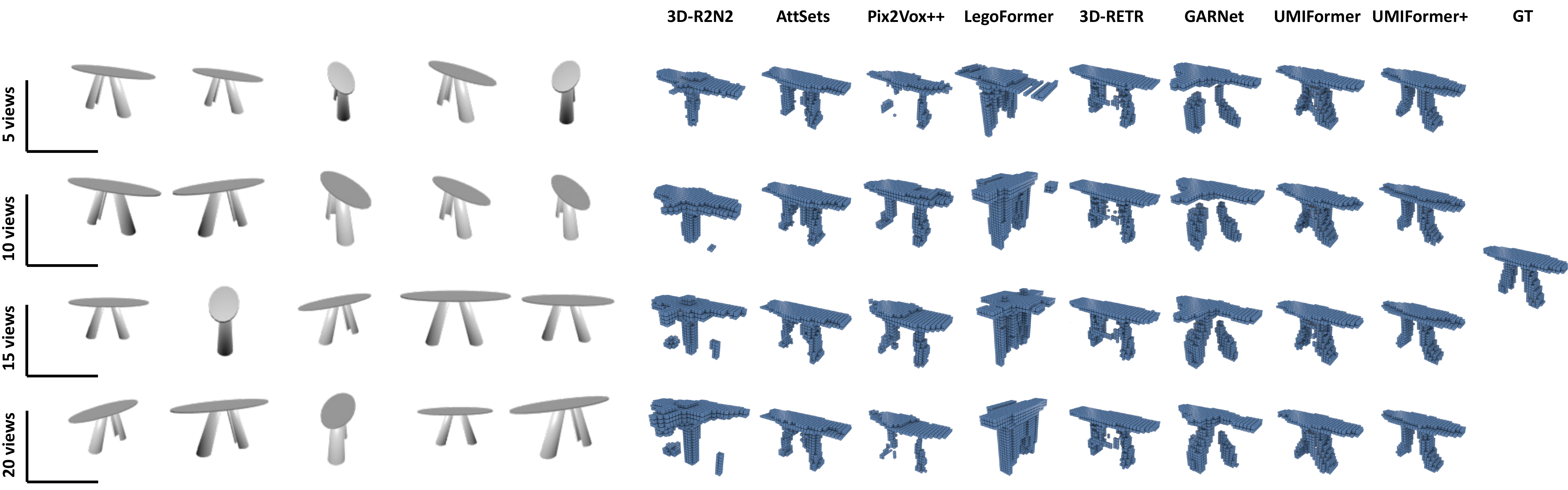}}\hspace{1000mm}
		\end{minipage}
		\\
		\begin{minipage}{1.03\linewidth}
			\centerline{\includegraphics[width=1\linewidth]{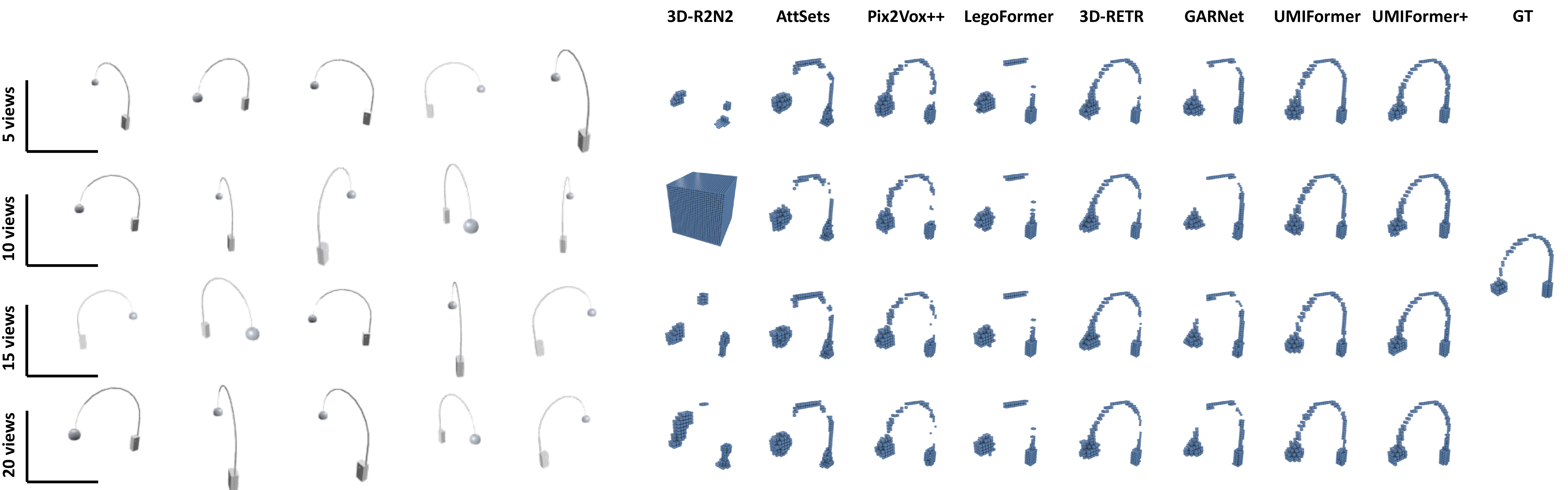}}\hspace{1000mm}
		\end{minipage}
		\\
		\begin{minipage}{1.03\linewidth}
			\centerline{\includegraphics[width=1\linewidth]{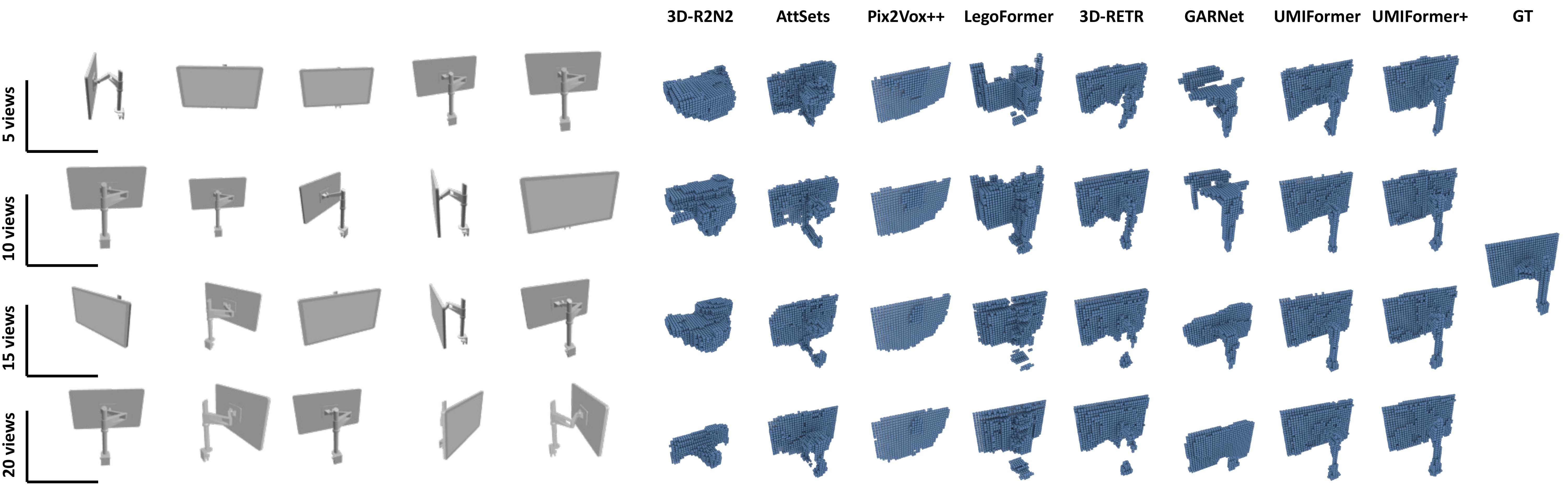}}
		\end{minipage}
		\caption{Qualitative reconstruction results when facing 5 views, 10 views, 15 views and 20 views as input for table, lamp and display.}
		\label{show_results_2}
	\end{figure*}

\begin{figure*}[ht]
		\centering
		\begin{minipage}{1.03\linewidth}
			\centerline{\includegraphics[width=1\linewidth]{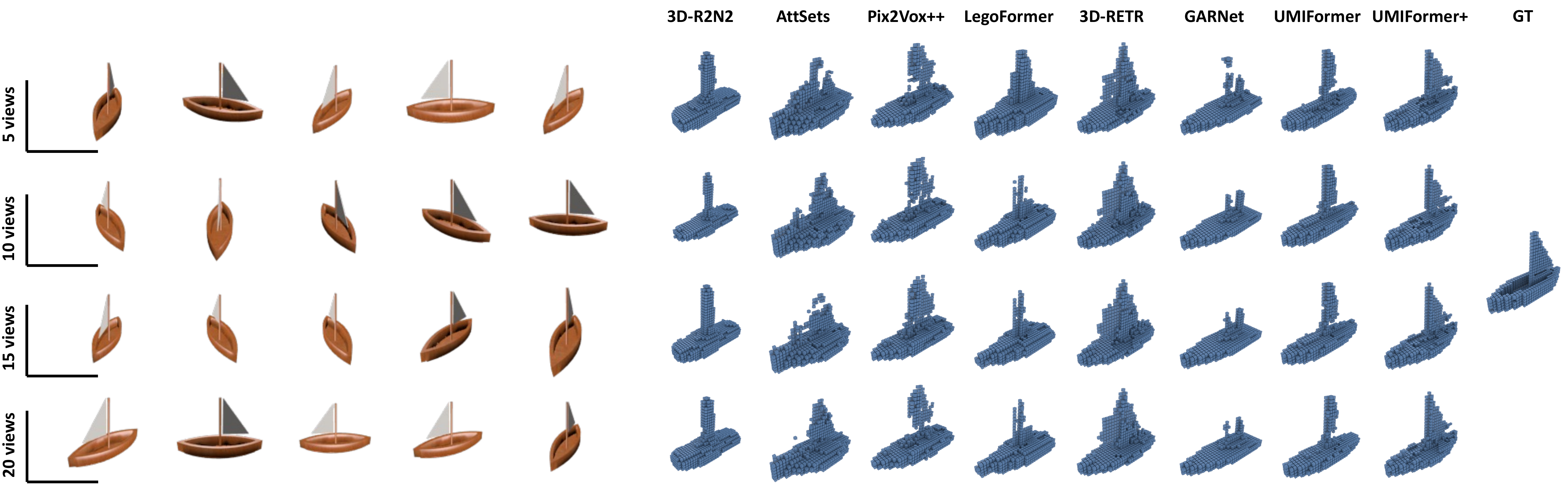}}\hspace{1000mm}
		\end{minipage}
		\\
		\begin{minipage}{1.03\linewidth}
			\centerline{\includegraphics[width=1\linewidth]{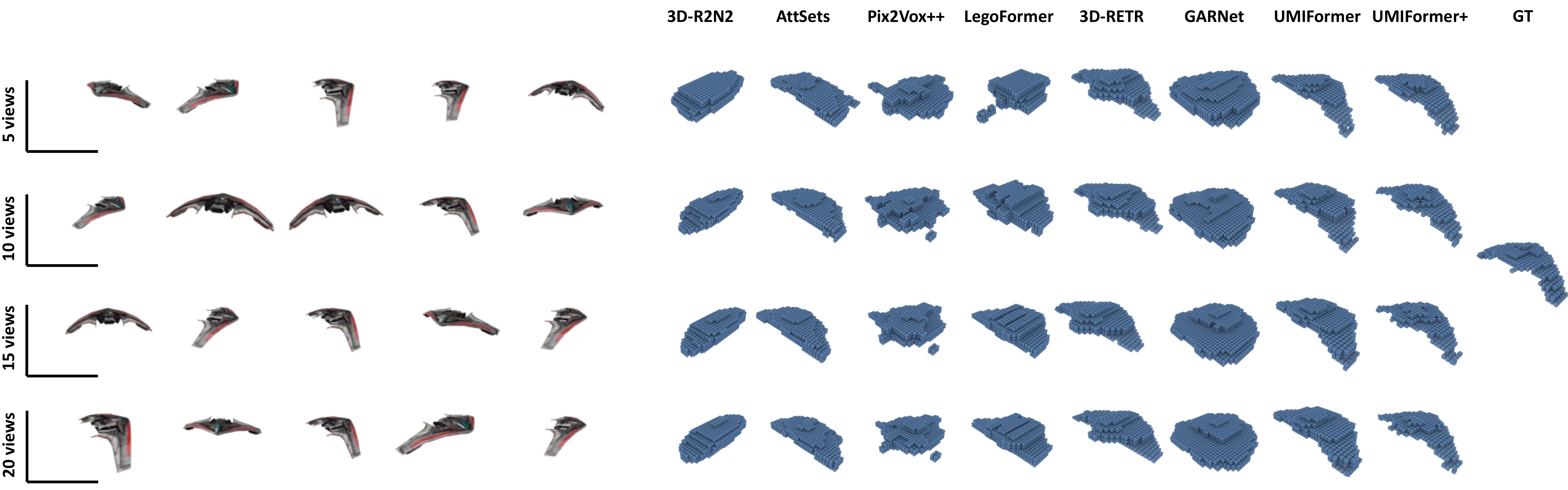}}\hspace{1000mm}
		\end{minipage}
		\\
		\begin{minipage}{1.03\linewidth}
			\centerline{\includegraphics[width=1\linewidth]{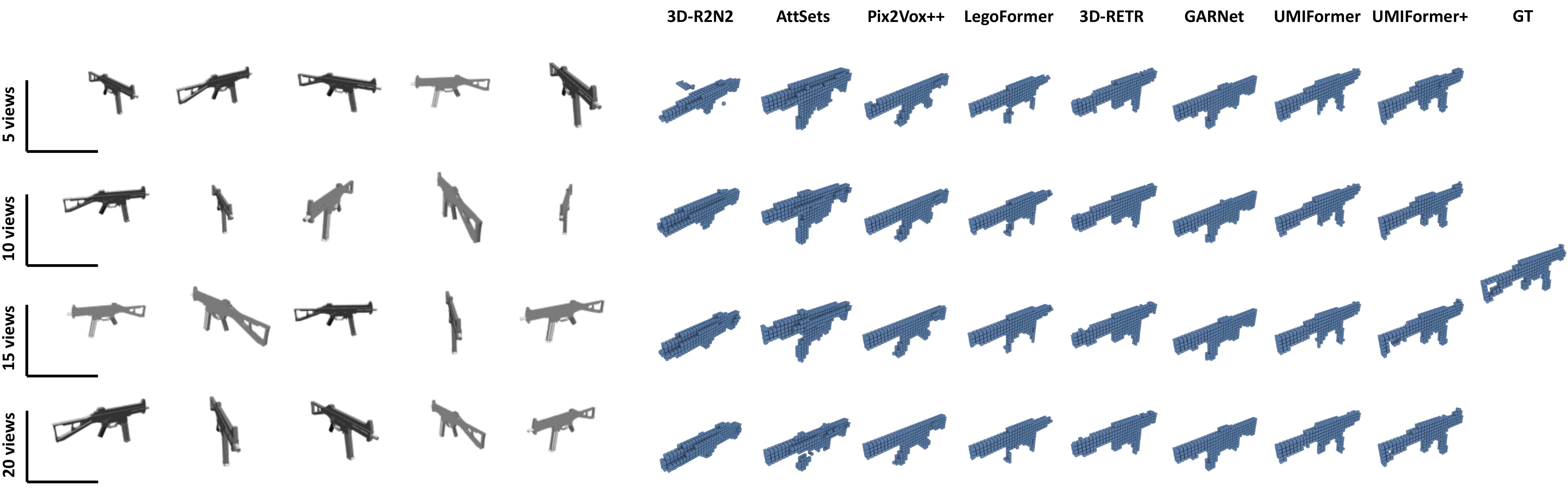}}
		\end{minipage}
		\caption{Qualitative reconstruction results when facing 5 views, 10 views, 15 views and 20 views as input for watercraft, airplane and rifle.}
		\label{show_results_3}
	\end{figure*}

	Figure~\ref{show_results_1}, Figure~\ref{show_results_2} and Figure~\ref{show_results_3} supplement more reconstruction examples on the test set of ShapeNet using various methods, including UMIFormer, UMIFormer+ and \cite{choy20163d, yang2020robust, xie2020pix2vox++, yagubbayli2021legoformer, shi20213d, zhu2023garnet} when facing 5 views, 10 views, 15 views and 20 views as input.
	
	\subsection{Various Decoders}
	
	In Table~\ref{different_decoder}, we provide extra ablation experiments that use other decoder architectures (from EVolT \cite{wang2021multi} and LegoFormer \cite{yagubbayli2021legoformer}) to supplement the experiment results in Table~\ref{ablation_experiments}. It verifies that IVDB and STM hold effective under various decoder networks. Among them, the performance when using the decoder of EVolT is even better than our proposed model shown in the main paper which uses the decoder of 3D-RETR \cite{shi20213d}. However, EVolT lacks some implementation details in the public information, hence we are worried whether our reproduction follows the original work exactly. Therefore, we do not use this better-performance version in this paper.
	
	\begin{table*}[h]
		\centering
		\renewcommand\arraystretch{1.2}
		\scalebox{0.7}{
			\begin{tabular}{cc|cccccc|cccccc}
				\hline
				\multirow{2}{*}{\textbf{IVDB}} & \multirow{2}{*}{\textbf{Fusion}} & \multicolumn{6}{c|}{\textbf{Decoder of EVolT \cite{wang2021multi}}} & \multicolumn{6}{c}{\textbf{Decoder of LegoFormer \cite{yagubbayli2021legoformer}}} \\
				& & \multicolumn{1}{c}{\textbf{3 views}} & \multicolumn{1}{c}{\textbf{5 views}} & \multicolumn{1}{c}{\textbf{8 views}} & \multicolumn{1}{c}{\textbf{12 views}} & \multicolumn{1}{c}{\textbf{16 views}} & \multicolumn{1}{c|}{\textbf{20 views}} & \multicolumn{1}{c}{\textbf{3 views}} & \multicolumn{1}{c}{\textbf{5 views}} & \multicolumn{1}{c}{\textbf{8 views}} & \multicolumn{1}{c}{\textbf{12 views}} & \multicolumn{1}{c}{\textbf{16 views}} & \multicolumn{1}{c}{\textbf{20 views}} \\ \hline
				\XSolidBrush & \textbf{ABM} & 0.7399 & 0.7484 & 0.7523 & 0.7546 & 0.7561 & 0.7567 & 0.7325 & 0.7401 & 0.7443 & 0.7465 & 0.7480 & 0.7485 \\
				\XSolidBrush & \textbf{STM} & 0.7504 & 0.7594 & 0.7631 & 0.7649 & 0.7655 & 0.7655 & 0.7453 & 0.7532 & 0.7581 & 0.7596 & 0.7610 & 0.7615 \\
				\Checkmark & \textbf{ABM} & 0.7419 & 0.7509 & 0.7555 & 0.7577 & 0.7592 & 0.7598 & 0.7425 & 0.7512 & 0.7560 & 0.7588 & 0.7605 & 0.7609 \\
				\Checkmark & \textbf{STM} & \textbf{0.7536} & \textbf{0.7633} & \textbf{0.7676} & \textbf{0.7699} & \textbf{0.7710} & \textbf{0.7714} & \textbf{0.7490} & \textbf{0.7589} & \textbf{0.7642} & \textbf{0.7664} & \textbf{0.7681} & \textbf{0.7682} \\ \hline
		\end{tabular}}
		\caption{Supplementary ablation experiments about various decoders.}
		\label{different_decoder}
	\end{table*}
	
	\subsection{Scheme of Inserting IVDB}
	
	Our proposed UMIformer model uses multiple IVDBs. In order to prove the necessity of this scheme, Table~\ref{ivdb_scheme} compares the performance when using IVDB once and using it repeatedly. Obviously, using IVDB once is also effective while weaker than using it repeatedly. It makes us realize that the number of IVDBs used is actually a tradeoff between performance and efficiency.
	
	\begin{table}[h]
		\centering
		\renewcommand\arraystretch{1.4}
		\scalebox{0.68}{
			\begin{tabular}{c|cccccc}
				\multicolumn{1}{c|}{\textbf{Scheme}} & \multicolumn{1}{c}{\textbf{3 views}} & \multicolumn{1}{c}{\textbf{5 views}} & \multicolumn{1}{c}{\textbf{8 views}} & \multicolumn{1}{c}{\textbf{12 views}} & \multicolumn{1}{c}{\textbf{16 views}} & \multicolumn{1}{c}{\textbf{20 views}} \\ \hline
				\textbf{w/o IVDB} & 0.7477 & 0.7557 & 0.7587 & 0.7598 & 0.7606 & 0.7606 \\
				\textbf{only IVDB once} & 0.7476 & 0.7564 & 0.7606 & 0.7624 & 0.7633 & 0.7636 \\
				\textbf{IVDB repeatedly} & \textbf{0.7518} & \textbf{0.7612} & \textbf{0.7661} & \textbf{0.7682} & \textbf{0.7696} & \textbf{0.7702} \\ \hline
		\end{tabular}}
		\caption{Comparison of the performance when using different schemes of inserting IVDB.}
		\label{ivdb_scheme}
	\end{table}
	
	\subsection{Pre-Training of ViT}
	
	In Section~\ref{sec:experiments:results}, we mention that inserting IVDB needs to preserve the pre-training advantages of ViT. Because the model performance relies heavily on it. If intra-view modules are without pre-learned parameters, the performance will be extremely poor as shown in Table~\ref{pre-training}. There is even an abnormal result that performance seriously degrades when the view increases.
	
	\begin{table}[h]
		\centering
		\renewcommand\arraystretch{1.5}
		\scalebox{0.68}{
			\begin{tabular}{c|cccccc}
				\multicolumn{1}{c|}{\textbf{Pre-training}} & \multicolumn{1}{c}{\textbf{3 views}} & \multicolumn{1}{c}{\textbf{5 views}} & \multicolumn{1}{c}{\textbf{8 views}} & \multicolumn{1}{c}{\textbf{12 views}} & \multicolumn{1}{c}{\textbf{16 views}} & \multicolumn{1}{c}{\textbf{20 views}} \\ \hline
				\XSolidBrush & 0.6075 & 0.5903 & 0.5505 & 0.5247 & 0.5144 & 0.5082 \\
				\Checkmark & \textbf{0.7518} & \textbf{0.7612} & \textbf{0.7661} & \textbf{0.7682} & \textbf{0.7696} & \textbf{0.7702} \\ \hline
		\end{tabular}}
		\caption{Comparison of the performance of whether the intra-view-decoupled transformer blocks are initialized by the pre-trained ViT.}
		\label{pre-training}
	\end{table}

\end{document}